\documentclass[journal]{IEEEtran}
\usepackage{amsmath,amsfonts}

\usepackage{amssymb}
\usepackage{mathtools}
\usepackage{amsthm}
\usepackage{algorithmic}
\usepackage{algorithm}
\usepackage{array}
\usepackage{hyperref}
\usepackage[caption=false,font=normalsize,labelfont=sf,textfont=sf]{subfig}
\usepackage{textcomp}
\usepackage{stfloats}
\usepackage{url}
\usepackage{verbatim}
\usepackage{graphicx}
\usepackage{multirow}
\usepackage{xcolor}
\usepackage[capitalize,noabbrev]{cleveref}

\usepackage{cite}
\usepackage{booktabs} 
\theoremstyle{plain}

\theoremstyle{definition}

\theoremstyle{remark}

\usepackage{adjustbox}
\hypersetup{
    colorlinks=true,
    linkcolor=blue,
    urlcolor=blue,
    citecolor=blue
}

\begin{document}

\title{AL-GNN: Replay-Free Analytic Learning for Class-Incremental Graph Learning}

\author{

Xuling Zhang, Jindong Li, Yifei Zhang, Mingqi Yang, Menglin Yang

\thanks{
Xuling Zhang, Jindong Li, and Menglin Yang are with The Hong Kong University of Science and Technology (Guangzhou), Guangzhou, China. Yifei Zhang is with the Northwestern Polytechnical University, Xian, China. Mingqi Yang is with the South China University of Technology, Guangzhou, China. E-mail: xling.zhang@outlook.com; jli839@connect.hkust-gz.edu.cn; yifeiacc@gmail.com; yangmq@scut.edu.cn; menglin.yang@outlook.com.
}

\thanks{Menglin Yang is corresponding author.}

}

% The paper headers
\markboth{IEEE Transactions on Pattern Analysis and Machine Intelligence}%
{Zhang \MakeLowercase{\textit{et al.}}: AL-GNN: Replay-Free Analytic Learning for Class-Incremental Graph Learning}

\maketitle

\begin{abstract}
Class-incremental graph learning aims to enable graph neural networks to continuously incorporate newly arriving classes without severely degrading performance on previously learned ones. Existing continual graph learning methods, especially replay-based approaches, often rely on storing and revisiting historical graph data to alleviate catastrophic forgetting. Although effective, such strategies introduce additional storage overhead, repeated optimization cost, and limited compatibility with privacy-constrained deployment scenarios. In this work, we instantiate this problem in the class-incremental graph node classification setting and propose AL-GNN, a replay-free analytic continual learning framework. AL-GNN decouples representation learning from continual classifier adaptation: it first trains a graph encoder in the base stage, and then replaces the gradient-trained classifier with an analytic classifier in an expanded feature space. During the later incremental stages, the classifier is updated recursively in closed form using only the currently revealed labeled data and a regularized feature autocorrelation matrix that summarizes previously observed representations. This design eliminates the need for replay buffers during incremental learning and substantially reduces dependence on repeated backpropagation. Under the fixed-feature formulation adopted in this work, the recursive update is equivalent to the corresponding joint analytic solution over all seen sessions. Extensive experiments on six benchmark datasets show that AL-GNN achieves strong average performance, competitive or lower forgetting on most datasets, and substantially lower training time than representative continual graph learning baselines. For instance, it improves average performance by 10\% on CoraFull
and reduces forgetting by over 30\% on Reddit,
while also reducing training time by nearly 50\%
due to its backpropagation-free design.
\end{abstract}

\section{Introduction}

Graph-structured data are pervasive in a wide range of real-world applications, such as citation analysis, social networks, recommender systems, and e-commerce platforms~\cite{2021_AI_Survey_Graph-Learning=A-Survey,2020_TSIP_Survey_Graph-Representation-Learning=A-Survey,2024_CSUR_Survey_A-Survey-of-Graph-Neural-Networks-for-Social-Recommender-Systems,2020_TKDE_Survey_A-Survey-on-Knowledge-Graph-based-Recommender-Systems,2022_CSUR_Survey_Graph-Neural-Network-in-Recommender-Systems=A-Survey,2023_AI_Survey_Continual-Learning=A-Review-of-Techniques-Challenges-and-Future-Directions}. In these applications, graph neural networks (GNNs) have become a powerful tool for node classification and representation learning by jointly modeling node attributes and graph topology~\cite{2020_TNNLS_Survey_A-Conprehensieve-Survey-on-Graph-Neural-Networks,2024_Journal-of-Big-Data_Survey_A-Review-of-Graph-Neural-Networks=Concepts-Architectures-Techniques-Challenges-Datasets-Applications-and-Future-Directions}. However, most existing GNNs are developed under an offline assumption, where all classes and labeled observations are available before training begins. This assumption is often unrealistic in evolving environments, where new supervision signals and semantic categories arrive progressively over time.

\begin{figure}[t]
    \centering
    \includegraphics[width=0.48\textwidth]{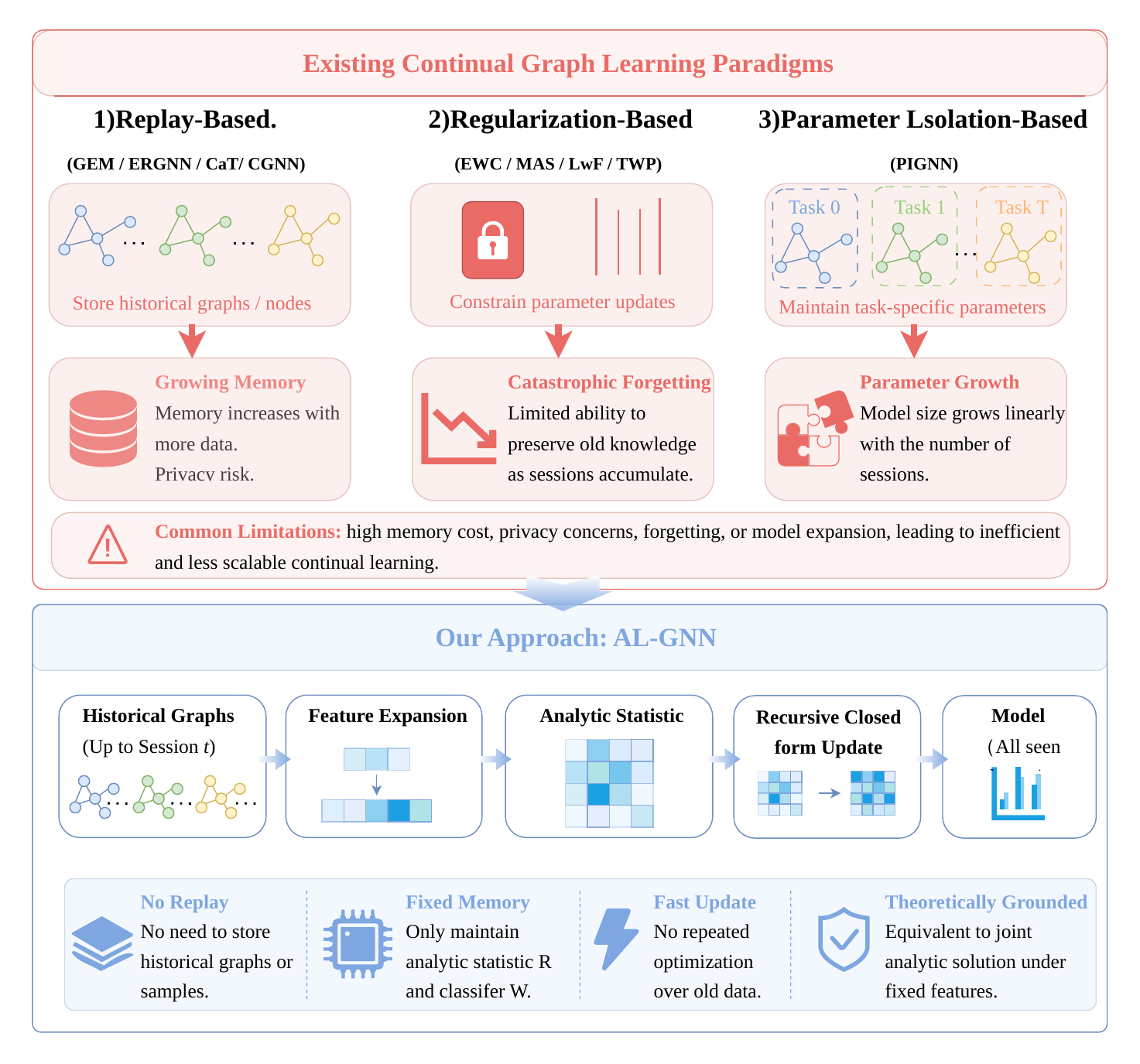}
    \caption{Challenges and motivation.}
    \label{fig:challenge-motivation}
\end{figure}

Such settings motivate continual graph learning (CGL), which studies how a graph model can be updated sequentially while preserving previously acquired knowledge~\cite{2023_arXiv_Survey_Continual-Graph-Learning=A-Survey,2024_arXiv_Survey_Continual-Learning-on-Graphs=Challenges-Solutions-and-Opportunities,2024_arXiv_Survey_Continual-Learning-on-Graphs=A-Survey}. Continual graph learning includes several protocols, such as class-incremental, task-incremental, and domain-incremental learning. In this paper, we focus on the class-incremental setting, where new classes are introduced session by session and the model is evaluated on all classes observed so far. In our study, this setting is instantiated on graph node classification benchmarks. This setting is practically important for graph-based systems that need to recognize emerging categories without retraining from scratch.

Class-incremental graph learning is challenging for at least three reasons. First, the label space expands over time, so the model must continually incorporate previously unseen semantic concepts while preserving discrimination among old classes. Second, historical graph data are often unavailable for long-term reuse due to storage limitations, privacy constraints, or deployment policies, making replay-based solutions undesirable in many scenarios. Third, graph representations depend on both node attributes and structural relationships; repeated gradient-based updates may therefore accumulate representation drift and exacerbate catastrophic forgetting over long incremental horizons. These factors make replay-free continual graph learning substantially more difficult than static offline graph learning.

Existing continual learning methods mainly address forgetting through replay, regularization, or parameter isolation, and these ideas have also been extended to graph neural networks. Replay-based methods are often strong empirical performers because historical samples or compressed graph memories provide direct access to earlier knowledge during later stages. However, this advantage comes at a cost: storing historical graph data increases memory consumption, replay introduces repeated optimization overhead, and retaining past graph samples may be problematic in privacy-sensitive applications. In contrast, non-replay methods avoid explicit memory buffers, but their retention ability often remains limited when the number of incremental sessions becomes large. Therefore, a key open problem is how to achieve replay-free class-incremental graph learning while preserving knowledge from previous sessions and maintaining computational efficiency.

To address this problem, we propose AL-GNN, a replay-free analytic continual learning framework for class-incremental graph learning. The core idea is to decouple representation learning from continual classifier adaptation. AL-GNN first trains a graph encoder in the base stage using standard backpropagation, thereby constructing a stable representation space from the initially available classes. It then replaces the gradient-trained classifier with an analytic classifier in an expanded feature space, whose parameters can be computed in closed form. During the later incremental stages, the encoder remains fixed and the classifier is updated recursively using only the currently revealed labeled data together with a regularized feature autocorrelation matrix accumulated from earlier stages. As a result, AL-GNN avoids storing historical node samples, eliminates repeated multi-epoch optimization of the classifier during incremental learning, and remains naturally compatible with privacy-constrained deployment settings.

The proposed analytic formulation offers several benefits. From an optimization perspective, it replaces repeated classifier retraining with recursive least-squares style updates, thereby substantially reducing the cost of incremental adaptation. From a memory perspective, it stores only fixed-size accumulated statistics rather than replay buffers or historical graph exemplars. From a continual learning perspective, under the fixed-feature formulation adopted in this work, the recursive update is equivalent to the corresponding joint analytic solution over all seen sessions, which provides a principled explanation for its strong retention behavior.

We evaluate AL-GNN on six graph node benchmarks under class-incremental settings. Experimental results show that AL-GNN achieves strong average performance, competitive or lower forgetting on most datasets, and substantially reduced training time compared with representative continual graph learning baselines. Moreover, AL-GNN remains robust under long and highly fragmented incremental horizons, where many competing methods become unstable.

The main contributions of this work are summarized as follows:
\begin{itemize}
    \item We propose AL-GNN, a replay-free analytic continual learning framework for class-incremental graph learning, which combines a pretrained graph encoder, an analytic classifier, and recursive closed-form updates.
    \item We derive a recursive analytic update rule that enables classifier adaptation using only current-session data and fixed-size accumulated statistics, without storing historical graph samples. 
    \item We conduct extensive experiments on six benchmark datasets and show that AL-GNN achieves a favorable trade-off among predictive performance, forgetting mitigation, efficiency, and long-horizon stability.
\end{itemize}

\section{Related Work}

\subsection{Continual Graph Learning}

Continual graph learning extends continual learning to graph-structured data and studies how graph neural networks can be updated sequentially without severely degrading performance on previously learned knowledge~\cite{2023_arXiv_Survey_Continual-Graph-Learning=A-Survey,2024_arXiv_Survey_Continual-Learning-on-Graphs=Challenges-Solutions-and-Opportunities,2024_arXiv_Survey_Continual-Learning-on-Graphs=A-Survey,2021_TPAMI_Survey_A-Continual-Learning-Survey=Defying-Forgetting-in-Classification-Tasks,2024_TPAMI_Survey_A-Comprehensive-Survey-of-Continual-Learning=Theory-Method-and-Application,goodfellow2013empirical}. In this work, we focus on the class-incremental setting, where newly arriving classes must be incorporated without revisiting all historical graph data. Existing methods can be broadly divided into replay-based, regularization-based, and parameter-isolation approaches.

Replay-based methods alleviate forgetting by storing historical information and reusing it during later updates. In the general continual learning literature, representative examples include GEM~\cite{2017_NeurIPS_GEM_Gradient-Episodic-Memory-for-Continual-Learning}, Gradient Projection Memory~\cite{saha2021gradient}, deep generative replay~\cite{shin2017continual}, and experience replay~\cite{rolnick2019experience}. In graph learning, ERGNN~\cite{2021_AAAI_ER-GNN_OVercoming-Catastrophic-Forgetting-in-Graph-Neural-Networks-with-Experience-Replay} stores representative graph samples for rehearsal, while CaT~\cite{2023_ICDM_CaT_CaT=Balanced-Continual-Graph-Learning-with-Graph-Condensation} reduces replay cost through graph condensation. These methods are often effective, but they introduce persistent storage requirements and repeated optimization overhead.

Regularization-based methods avoid explicit replay by constraining parameter updates so that important knowledge is preserved during later training. Representative examples include EWC~\cite{2017_NAS_EWC_Overcoming-Catastrophic-Forgetting-in-Neural-Networks}, Synaptic Intelligence~\cite{pmlr-v70-zenke17a}, LwF~\cite{2017_TPAMI_LwF_Learning-without-Forgetting}, less-forgetting learning~\cite{jung2016less}, and MAS~\cite{2018_ECCV_MAS_Memory-Aware-Synapses}. In the graph domain, TWP~\cite{2021_AAAI_TWP_Overcoming-Catastrophic-Forgetting-in-Graph-Neural-Networks} introduces task-aware regularization tailored to continual GNN training, while benchmark studies have documented the severity of forgetting in graph settings~\cite{carta2021catastrophic}. Although these methods do not require replay buffers, they still rely on repeated backpropagation and often face a difficult stability--plasticity trade-off when the number of incremental sessions becomes large.

A third line of work relies on parameter isolation, decomposition, or model expansion. PIGNN~\cite{2023_SIGIR_PI-GNN_Continual-Learning-on-Dynamic-Graphs-via-Parameter-Isolation} reduces interference across tasks by allocating task-specific parameters or subnetworks, and additive parameter decomposition has also been shown to improve order robustness in continual learning~\cite{yoon2020scalable}. Hybrid frameworks such as ContinualGNN~\cite{2020_CIKM_ContinualGNN_Streaming-Graph-Neural-Networks-via-Continual-Learning} combine multiple mechanisms to improve retention. More broadly, continual graph learning should be distinguished from dynamic graph representation learning~\cite{2021_CSUR_Survey_A-Survey-on-Embedding-Dynamic-Graphs,2020_PR_Dynamic-Graph-Convolutional-Networks,2022_KDD_ROLAND_ROLAND=Graph-Learning-Framework-for-Dynamic-Grpahs,2022_ToC_A-Novel-Representation-Learning-for-Dynamic-Graphs-based-on-Graph-Convolutional-Networks}: the latter focuses on evolving graph structure, whereas the former focuses on sequential knowledge accumulation under forgetting constraints.

Compared with these approaches, AL-GNN follows a different design principle. Rather than storing historical graph data, constraining gradient-based updates, or expanding task-specific subnetworks, it maintains a fixed-size recursive statistic in the representation space and updates the classifier through closed-form analytic recursion.

\subsection{Analytic Learning and Recursive Least-Squares Updates}

Analytic learning offers an alternative to standard backpropagation by estimating model parameters through closed-form or recursive optimization, typically under least-squares objectives~\cite{zhuang2024dsaldualstreamanalyticlearning}. Early analytic models, including radial basis function networks~\cite{1988_Complex-Systems_RBF_Multivariable-Functional-Interpolation-and-Adaptive-Networks} and extreme learning machines~\cite{2006_NeuroComputing_ELM_Extreme-Learning-Machine-Theory-and-Applications}, separate feature transformation from linear parameter estimation, thereby enabling efficient training without iterative gradient descent. Recursive matrix update techniques such as the Block-wise Recursive Moore--Penrose approach~\cite{2021_TSMC_BRMP_Blockwise-Recursive-Moore-Penrose-Inverse-for-Network-Learning} further show that analytic solutions can be updated incrementally without recomputing large inverse matrices from scratch.

Despite these advantages, analytic learning has been only lightly explored in continual graph learning. Most existing continual GNN methods still rely on backpropagation-based adaptation, either with replay or with parameter regularization. As a result, the potential of recursive analytic updates for replay-free graph continual learning remains underexplored. Our work addresses this gap by combining a graph encoder trained in the base stage with a recursively updated analytic classifier for later incremental stages. In this sense, AL-GNN is related to recursive least-squares learning, but differs from prior analytic methods in its explicit focus on class-incremental graph node classification and its design for replay-free continual adaptation.

\begin{figure*}[t]
    \centering
    \includegraphics[width=\textwidth]{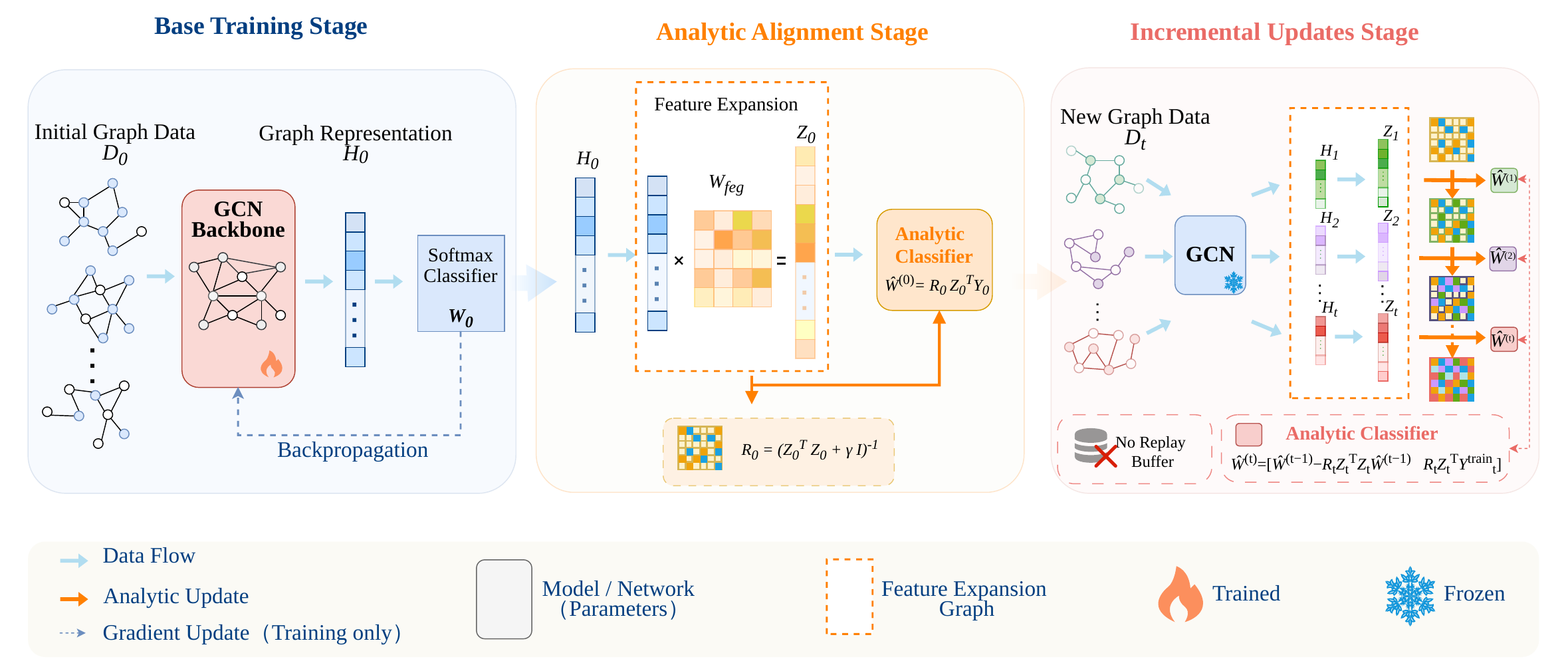}
    \caption{Overview of AL-GNN. The framework consists of three stages: (1) Base training stage, where the GCN backbone is trained on initial graph data $\mathcal{D}_0$ via backpropagation; (2) Analytic alignment stage, where the analytic classifier $\hat{\mathbf{W}}^{(0)}$ is initialized in closed form using expanded features $\mathbf{Z}_0$; (3) Incremental updates stage, where the frozen GCN encoder extracts features from new graph data $\mathcal{D}_t$, followed by recursive analytic updates of the classifier without storing any replay buffer.}
    \label{fig:framework-al}
\end{figure*}

\section{Methodology}

\subsection{Problem Formulation}

We consider class-incremental node classification on a graph observed over a sequence of sessions. Let $\mathbf{A}$ denote the graph adjacency matrix and $\mathbf{X} \in \mathbb{R}^{N \times d}$ the node feature matrix, where $N$ is the number of nodes and $d$ is the input feature dimension. Each labeled node belongs to a class in the label space $\mathcal{Y}=\{1,2,\dots,C\}$. Continual graph learning includes several protocols, such as class-incremental, task-incremental, and domain-incremental learning. In this work, we focus on the class-incremental setting, where newly arriving sessions introduce previously unseen classes and the model must retain performance on all classes observed so far. Following the standard class-incremental node classification protocol used by representative continual graph learning baselines such as ERGNN, CaT, EWC, PIGNN, TWP, and LwF, the graph structure and node attributes remain available throughout training, while labeled supervision is revealed session by session according to the class split.

Specifically, the label space $\mathcal{Y}$ is divided into a base label set and a sequence of incremental label sets. The first $C_0=\lceil C/2 \rceil$ classes form the base session, denoted by $\mathcal{Y}_0=\{1,2,\dots,C_0\}$. The remaining classes are introduced sequentially in later sessions, denoted by $\mathcal{Y}_1,\mathcal{Y}_2,\dots,\mathcal{Y}_T$, where $\mathcal{Y}_t \cap \mathcal{Y}_s = \emptyset$ for $t \neq s$ and
\begin{equation}
\mathcal{Y} = \bigcup_{t=0}^{T} \mathcal{Y}_t.
\end{equation}
For simplicity, in our main experiments each incremental session introduces one new class, although the proposed framework can be extended to multi-class increments. Throughout the paper, the basic sequential unit is a session; a session may contain one or more newly introduced classes, and the one-class-per-session setting is emphasized because it is the most challenging case.

Accordingly, the labeled training nodes arrive as a sequence of sessions
\begin{equation}
\mathcal{D}^{\mathrm{train}} = \{\mathcal{D}_0^{\mathrm{train}}, \mathcal{D}_1^{\mathrm{train}}, \dots, \mathcal{D}_T^{\mathrm{train}}\},
\end{equation}
where $\mathcal{D}_0^{\mathrm{train}}$ contains nodes from the base classes $\mathcal{Y}_0$, and each $\mathcal{D}_t^{\mathrm{train}}$ for $t \geq 1$ contains only nodes from the newly introduced classes in session $t$. After finishing session $t$, the model is evaluated on test nodes drawn from all classes observed so far, i.e., from $\bigcup_{s=0}^{t} \mathcal{Y}_s$.

The goal is to learn a sequence of node classifiers $\{f_t(\cdot)\}_{t=0}^{T}$ such that, after training on the current session $\mathcal{D}_t^{\mathrm{train}}$, the model maintains strong predictive performance on both newly introduced and previously learned classes. Importantly, during the update from session $t-1$ to session $t$, the learner is not allowed to revisit historical training nodes $\{\mathcal{D}_0^{\mathrm{train}},\dots,\mathcal{D}_{t-1}^{\mathrm{train}}\}$. This restriction reflects realistic deployment scenarios in which past graph data cannot be stored or replayed due to privacy, storage, or system constraints.

Under this protocol, the main challenge is to update the model using only current-session supervision while preserving previously acquired knowledge. Our objective is therefore to design a replay-free continual learning framework for class-incremental graph learning that remains computationally efficient, avoids explicit storage of historical samples, and mitigates catastrophic forgetting throughout the incremental process.

In the setting considered in this paper, the graph structure and node attributes follow the same protocol as in the representative continual graph learning baselines considered in our experiments: they remain available to the encoder at each session, while the labeled supervision used for classifier adaptation is revealed incrementally according to the session split. Evaluation after session $t$ is always performed in the cumulative class space $\bigcup_{s=0}^{t}\mathcal{Y}_s$, so the learner must preserve discrimination over all seen classes rather than over the current session alone. This protocol is deliberately stringent: it removes replay while still requiring a single unified classifier over the expanding label space, which makes forgetting control substantially more difficult than in session-isolated settings.

\subsection{Overview of AL-GNN}

We now describe AL-GNN under the class-incremental setting considered in our experiments. The key design principle is to decouple representation learning from continual classifier adaptation. Rather than repeatedly fine-tuning the entire graph neural network as new classes arrive, AL-GNN first learns a graph encoder in the base stage and then performs replay-free analytic classifier updates in a frozen representation space.

As illustrated in Fig.~\ref{fig:framework-al}, the framework consists of three stages. First, a GCN encoder is trained in the base stage to construct a stable node representation space. Second, the stage-1 representations are mapped into an expanded feature space, where the initial classifier is obtained by a regularized closed-form solution together with initial analytic statistics. Third, during the later incremental stages, the encoder remains fixed and only the analytic classifier is updated using the newly revealed labeled data and the accumulated statistics from previous stages.

This design yields a compact and coherent continual learning pipeline. The base stage is responsible for representation learning and analytic initialization, whereas the later incremental stages are handled entirely by recursive closed-form updates. Consequently, AL-GNN does not store historical node samples, avoids repeated multi-epoch optimization during later stages, and preserves the information required for continual adaptation through fixed-size analytic statistics. The next subsections detail the initialization, recursive update rule, theoretical justification, and algorithmic summary of the proposed framework.

\subsection{Base-Stage Analytic Initialization}

The base stage serves to initialize both the graph encoder and the analytic classifier. Let $\mathbf{A}_0$, $\mathbf{X}_0$, and $\mathbf{Y}_0^{\mathrm{train}}$ denote the adjacency matrix, node feature matrix, and label matrix of the base classes, respectively. Here $\mathbf{Y}_t^{\mathrm{train}}$ is represented as a one-hot label matrix associated with the newly introduced classes revealed at incremental step $t$, and $\mathbf{Y}_t^{\mathrm{train}} \in \mathbb{R}^{n_t \times |\mathcal{Y}_t|}$, where $n_t$ is the number of labeled training nodes at step $t$. The cumulative number of seen classes after step $t$ is denoted by $C_t^{\mathrm{seen}}=\left|\bigcup_{s=0}^{t}\mathcal{Y}_s\right|$. AL-GNN first trains a GCN backbone on the base classes using standard backpropagation. This step is used only to learn a stable node representation space from the initially available classes. After this stage, the encoder parameters are fixed and reused throughout all later incremental stages. Accordingly, the predictor after incremental step $t$ is implemented by the frozen encoder, the fixed feature expansion layer, and the analytic classifier $\hat{\mathbf{W}}_{\mathrm{cls}}^{(t)}$.

Given the trained encoder, the base-stage node representations are obtained as
\begin{equation}
\mathbf{H}_0 = f_{\mathrm{gcn}}(\mathbf{X}_0,\mathbf{A}_0;\mathbf{W}_{\mathrm{GCN}}),
\end{equation}
where $\mathbf{H}_0 \in \mathbb{R}^{N_0 \times d_{\mathrm{gcn}}}$ denotes the node embedding matrix of the base stage, and $\mathbf{W}_{\mathrm{GCN}}$ denotes the learned encoder parameters. To enhance the expressive capacity of the analytic classifier, we further map these representations into an expanded feature space through a fixed feature expansion layer:
\begin{equation}
\mathbf{Z}_0 = \sigma(\mathbf{H}_0\mathbf{W}_{\mathrm{feg}}),
\end{equation}
where $\mathbf{W}_{\mathrm{feg}} \in \mathbb{R}^{d_{\mathrm{gcn}} \times d_{\mathrm{feg}}}$ is randomly initialized and fixed, $d_{\mathrm{feg}} > d_{\mathrm{gcn}}$, and $\sigma(\cdot)$ denotes an element-wise nonlinearity. This expansion serves two purposes. First, it increases the expressive capacity of the analytic classifier by mapping the learned GCN representations into a richer nonlinear feature space. Second, it preserves the simplicity of closed-form linear estimation, because the adaptive component remains the classifier in the expanded space rather than the encoder itself. In this sense, the feature expansion layer plays a role analogous to random feature mappings in analytic learning: it enriches the representation available to the classifier while keeping the subsequent update rule amenable to efficient recursive optimization.

Based on the expanded features $\mathbf{Z}_0$, the classifier is initialized by solving a regularized least-squares problem:
\begin{equation}
\hat{\mathbf{W}}_{\mathrm{cls}}^{(0)}
=
\arg\min_{\mathbf{W}}
\left\|\mathbf{Y}_0^{\mathrm{train}}-\mathbf{Z}_0\mathbf{W}\right\|_F^2
+\gamma\left\|\mathbf{W}\right\|_F^2,
\label{eq:base_ridge}
\end{equation}
where $\gamma > 0$ is a regularization coefficient. The closed-form solution is given by
\begin{equation}
\hat{\mathbf{W}}_{\mathrm{cls}}^{(0)}
=
(\mathbf{Z}_0^{\top}\mathbf{Z}_0+\gamma\mathbf{I})^{-1}\mathbf{Z}_0^{\top}\mathbf{Y}_0^{\mathrm{train}}.
\label{eq:base_closed_form}
\end{equation}
This initialization replaces the original gradient-trained classifier with an analytic classifier defined in the expanded feature space. The detailed derivation of Eq.~\eqref{eq:base_closed_form} follows standard multi-output ridge regression and is provided in the supplementary material.

In addition to the classifier weights, the base stage also initializes the regularized feature autocorrelation matrix,
\begin{equation}
\mathbf{R}_0=(\mathbf{Z}_0^{\top}\mathbf{Z}_0+\gamma\mathbf{I})^{-1},
\label{eq:R0}
\end{equation}
which serves as the analytic statistics for subsequent recursive updates. Therefore, after the base stage, AL-GNN no longer needs to optimize the classifier via iterative backpropagation; instead, it carries forward the frozen encoder, the initial analytic classifier $\hat{\mathbf{W}}_{\mathrm{cls}}^{(0)}$, and the accumulated statistics $\mathbf{R}_0$ into the later incremental stages.

\subsection{Replay-Free Recursive Analytic Update}

After the base stage has initialized the frozen encoder, the analytic classifier, and the analytic statistics, AL-GNN proceeds to the later incremental stages. At incremental step $t \geq 1$, the expanded features of the newly revealed labeled nodes are obtained by reusing the frozen encoder and fixed feature expansion mapping:
\begin{equation}
\mathbf{Z}_t = \sigma\bigl(f_{\mathrm{gcn}}(\mathbf{X}_t,\mathbf{A}_t;\mathbf{W}_{\mathrm{GCN}})\mathbf{W}_{\mathrm{feg}}\bigr).
\end{equation}
Under this fixed-feature setting, the ideal classifier after observing sessions $0$ through $t$ is the joint regularized least-squares solution in the cumulatively expanded output space. Let
\begin{equation}
\mathbf{Q}_t=
\left[
\mathbf{Z}_0^{\top}\mathbf{Y}_0^{\mathrm{train}}
\;\;
\mathbf{Z}_1^{\top}\mathbf{Y}_1^{\mathrm{train}}
\;\; \cdots \;\;
\mathbf{Z}_t^{\top}\mathbf{Y}_t^{\mathrm{train}}
\right]
\label{eq:Qt_def}
\end{equation}
denote the concatenated feature--label correlation matrix over all seen sessions. Then the joint analytic classifier after session $t$ can be written as
\begin{equation}
\hat{\mathbf{W}}_{\mathrm{cls}}^{(t)}=
\left(\sum_{s=0}^{t}\mathbf{Z}_s^{\top}\mathbf{Z}_s+\gamma\mathbf{I}\right)^{-1}
\mathbf{Q}_t.
\label{eq:joint_closed_form}
\end{equation}
Equivalently, this solution is obtained by solving the joint regularized least-squares problem over the cumulatively expanded classifier matrix. Directly evaluating Eq.~\eqref{eq:joint_closed_form} would require access to all historical expanded features, which is incompatible with replay-free continual learning.

To avoid this dependence, AL-GNN maintains the regularized feature autocorrelation inverse
\begin{equation}
\mathbf{R}_t=
\left(\sum_{s=0}^{t}\mathbf{Z}_s^{\top}\mathbf{Z}_s+\gamma\mathbf{I}\right)^{-1},
\label{eq:Rt_def}
\end{equation}
which compactly summarizes the second-order information of all seen sessions. Using Eq.~\eqref{eq:Rt_def}, the statistic can be updated recursively as
\begin{equation}
\mathbf{R}_t=
\left(\mathbf{R}_{t-1}^{-1}+\mathbf{Z}_t^{\top}\mathbf{Z}_t\right)^{-1}.
\label{eq:Rt_recursive}
\end{equation}
For efficient implementation, we further apply the Woodbury matrix identity and obtain
\begin{equation}
\mathbf{R}_t
=
\mathbf{R}_{t-1}
-\mathbf{R}_{t-1}\mathbf{Z}_t^{\top}
\left(\mathbf{I}+\mathbf{Z}_t\mathbf{R}_{t-1}\mathbf{Z}_t^{\top}\right)^{-1}
\mathbf{Z}_t\mathbf{R}_{t-1}.
\label{eq:Rt_woodbury}
\end{equation}
To write the recursive classifier update in the cumulatively expanded output space, we follow the block-concatenation form used in the supplementary derivation. The classifier is updated using only the current-session features, labels, and the stored statistic:
\begin{equation}
\hat{\mathbf{W}}_{\mathrm{cls}}^{(t)}
=
\left[
\hat{\mathbf{W}}_{\mathrm{cls}}^{(t-1)}
-\mathbf{R}_t\mathbf{Z}_t^{\top}\mathbf{Z}_t\hat{\mathbf{W}}_{\mathrm{cls}}^{(t-1)}
\quad
\mathbf{R}_t\mathbf{Z}_t^{\top}\mathbf{Y}_t^{\mathrm{train}}
\right].
\label{eq:Wt_recursive}
\end{equation}
Here $\mathbf{Y}_t^{\mathrm{train}}$ contains one-hot targets only for the classes newly introduced at incremental step $t$, so the right-hand block appends the classifier parameters associated with the new class columns, while the left-hand block recursively corrects the previously learned classifier. Accordingly, Eq.~\eqref{eq:Wt_recursive} updates a single cumulative classifier through matrix concatenation rather than through same-width addition. The detailed derivation is provided in the supplementary material.

Therefore, each later incremental stage is handled by propagating the pair $\{\hat{\mathbf{W}}_{\mathrm{cls}}^{(t-1)},\mathbf{R}_{t-1}\}$, extracting the current expanded features, and applying Eqs.~\eqref{eq:Rt_woodbury} and \eqref{eq:Wt_recursive}. This update rule removes the need for replay buffers while preserving the recursive analytic statistics required for continual adaptation.

\subsection{Theoretical Analysis of Replay-Free Analytic Updates}

The central theoretical question behind AL-GNN is whether the replay-free recursion preserves the same classifier that would have been obtained by jointly solving the analytic objective over all previously seen sessions. This question is answered affirmatively under the fixed-feature formulation used in this paper, where both $\mathbf{W}_{\mathrm{GCN}}$ and $\mathbf{W}_{\mathrm{feg}}$ remain unchanged after the base stage.

\paragraph{Equivalence to the joint analytic statistic.}
By definition, Eq.~\eqref{eq:Rt_def} is exactly the inverse Gram matrix appearing in the joint analytic solution in Eq.~\eqref{eq:joint_closed_form}. Eq.~\eqref{eq:Rt_recursive} therefore follows directly by separating the contribution of the current incremental step from the accumulated contributions of previous steps. Moreover, the Woodbury expansion in Eq.~\eqref{eq:Rt_woodbury} shows that this statistic can be updated from the previously stored matrix and the current-step features only.

\paragraph{Equivalence to the joint analytic classifier.}
Let
\begin{equation}
\mathbf{Q}_t=
\left[
\mathbf{Z}_0^{\top}\mathbf{Y}_0^{\mathrm{train}}
\;\;
\mathbf{Z}_1^{\top}\mathbf{Y}_1^{\mathrm{train}}
\;\; \cdots \;\;
\mathbf{Z}_t^{\top}\mathbf{Y}_t^{\mathrm{train}}
\right]
\end{equation}
denote the concatenated feature--label correlation matrix over all seen sessions. Then Eq.~\eqref{eq:joint_closed_form} can be written compactly as
\begin{equation}
\hat{\mathbf{W}}_{\mathrm{cls}}^{(t)} = \mathbf{R}_t\mathbf{Q}_t.
\label{eq:joint_compact}
\end{equation}
Because $\mathbf{Q}_t = [\mathbf{Q}_{t-1} \; \mathbf{Z}_t^{\top}\mathbf{Y}_t^{\mathrm{train}}]$ and $\hat{\mathbf{W}}_{\mathrm{cls}}^{(t-1)}=\mathbf{R}_{t-1}\mathbf{Q}_{t-1}$, substituting Eqs.~\eqref{eq:Rt_woodbury} and \eqref{eq:joint_compact} yields Eq.~\eqref{eq:Wt_recursive}. Hence, after every incremental step, the replay-free recursive update produces the same cumulative classifier as the joint analytic solution in Eq.~\eqref{eq:joint_closed_form} under the fixed-feature formulation.

\paragraph{Implication and scope.}
This equivalence explains why forgetting can be reduced without replay under the fixed-feature setting adopted here. At the same time, the result depends on the encoder and the expansion mapping remaining unchanged after the base stage. If the representation space changes across incremental steps, the exact equivalence no longer follows directly.

\subsection{Algorithmic Summary and Practical Properties}

Algorithm~\ref{alg:algnn} summarizes the complete training pipeline of AL-GNN. After the base stage, the method carries forward only the classifier parameters together with the analytic statistics defined in Eqs.~\eqref{eq:R0} and \eqref{eq:Rt_woodbury}; these maintained quantities are used in the later incremental stages.

\paragraph{Computational complexity.}
After the base stage, AL-GNN freezes the graph encoder and updates only the analytic classifier in the expanded feature space. Let $R=d_{\mathrm{feg}}^2$ denote the size of the analytic memory matrix. The dominant update cost therefore scales as $\mathcal{O}(R)$ per incremental session. Using Eq.~\eqref{eq:Rt_woodbury}, the recursive update is computed only from the current-session statistics $\mathbf{I}+\mathbf{Z}_t\mathbf{R}_{t-1}\mathbf{Z}_t^{\top}$ rather than from recomputation over all historical data. Moreover, once the analytic memory matrix $\mathbf{R}_t$ is formed, the updates in Eq.~\eqref{eq:Rt_woodbury} and Eq.~\eqref{eq:Wt_recursive} can be efficiently implemented using parallel GPU matrix operations, resulting in low practical incremental overhead.

% ---------------------------- pseudo code -------------------------- %
\begin{table}[t]
\centering
\footnotesize
\caption{Algorithmic summary of AL-GNN.}
\label{alg:algnn}

\renewcommand{\arraystretch}{1.08}
\begin{tabular}{p{0.96\linewidth}}
\toprule
\textbf{Algorithm 1: AL-GNN} \\
\midrule
\textbf{Input:} $\{\mathcal{D}_t^{\mathrm{train}}\}_{t=0}^{T}$, $\mathbf{A}$, $\mathbf{X}$, $f_{\mathrm{gcn}}$, $\mathbf{W}_{\mathrm{feg}}$, $\gamma$ \\
\textbf{Output:} $\hat{\mathbf{W}}_{\mathrm{cls}}^{(T)}$ \\[0.4ex]

\textbf{Stage 1: Base representation learning} \\
1: Train $f_{\mathrm{gcn}}$ on $\mathcal{D}_0^{\mathrm{train}}$ \\
2: Compute $\mathbf{H}_0 \leftarrow f_{\mathrm{gcn}}(\mathbf{X}_0,\mathbf{A}_0;\mathbf{W}_{\mathrm{GCN}})$ \\
3: Freeze $\mathbf{W}_{\mathrm{GCN}}$ \\[0.6ex]

\textbf{Stage 2: Analytic alignment} \\
4: Compute $\mathbf{Z}_0 \leftarrow \sigma(\mathbf{H}_0\mathbf{W}_{\mathrm{feg}})$ \\
5: Initialize $\hat{\mathbf{W}}_{\mathrm{cls}}^{(0)}$ by Eq.~\eqref{eq:base_closed_form} \\
6: Initialize $\mathbf{R}_0$ by Eq.~\eqref{eq:R0} \\
7: Keep $\mathbf{W}_{\mathrm{feg}}$ fixed \\[0.6ex]

\textbf{Stage 3: Incremental analytic updates} \\
8: \textbf{for} $t=1,2,\ldots,T$ \textbf{do} \\
\quad 9: Compute $\mathbf{Z}_t \leftarrow \sigma(f_{\mathrm{gcn}}(\mathbf{X}_t,\mathbf{A}_t;\mathbf{W}_{\mathrm{GCN}})\mathbf{W}_{\mathrm{feg}})$ \\
\quad 10: Update $\mathbf{R}_t$ by Eq.~\eqref{eq:Rt_woodbury} \\
\quad 11: Update $\hat{\mathbf{W}}_{\mathrm{cls}}^{(t)}$ by Eq.~\eqref{eq:Wt_recursive} \\
\quad 12: Discard raw data of session $t$ and retain $\{\hat{\mathbf{W}}_{\mathrm{cls}}^{(t)},\mathbf{R}_t\}$ \\
13: \textbf{end for} \\
\textbf{return} $\hat{\mathbf{W}}_{\mathrm{cls}}^{(T)}$ \\
\bottomrule
\end{tabular}

\end{table}

Table~\ref{tab:complexity_comparison} summarizes the additional continual learning overhead beyond standard backbone pretraining. Here, the backbone parameter scale is represented by $P_b=d_{in}d_h+d_hd_{out}$, where $d_{in}$, $d_h$, and $d_{out}$ denote the input, hidden, and output embedding dimensions of the graph encoder, respectively. In commonly used graph continual learning benchmarks, $d_h$ is typically chosen from $128$ to $512$, whereas the feature expansion ratio in AL-GNN usually satisfies $k\leq10$.

Replay-based methods additionally depend on the retained replay graph size. In GEM, the replay memory stores historical replay nodes, resulting in a storage complexity of $\mathcal{O}(Md)$, where $M$ denotes the number of retained replay samples and $d$ denotes the stored node representation dimension. Since replay optimization additionally requires repeated gradient updates over replay batches, its practical training cost further scales as $\mathcal{O}(EMP_b)$, where $E$ denotes the number of optimization epochs.

ERGNN additionally stores replay graph structure information. Its complexity is therefore represented as $\mathcal{O}(Md+E_M)$, where $E_M$ denotes the number of retained replay edges or local graph connections associated with replay nodes. Since replay subgraphs usually expand together with retained historical nodes, the replay graph size increases continuously during continual learning. Similarly, CaT stores condensed replay graphs with complexity $\mathcal{O}(M_c(d+\bar{k}))$, where $M_c$ denotes the number of condensed replay nodes generated by graph condensation and $\bar{k}$ denotes the average retained graph connectivity. Although graph condensation reduces replay size, the retained replay graph still scales with historical graph information. CGNN similarly maintains replay graph information with complexity $\mathcal{O}(M(d+\bar{k}))$.

By contrast, the memory complexity of AL-GNN depends only on the analytic memory matrix size $R=d_{\mathrm{feg}}^2$, where $d_{\mathrm{feg}}=kd_h$. Although this introduces a quadratic dependency on the representation dimension, the feature expansion ratio $k$ remains a small constant after initialization and does not grow with continual learning. Consequently, the retained memory of AL-GNN remains bounded throughout incremental learning and is independent of the number of historical nodes, replay edges, or incremental sessions.

Another important distinction lies in the optimization procedure. Most replay-based and regularization-based continual learning methods require iterative gradient optimization over multiple epochs, leading to an additional multiplicative training factor $E$. In contrast, after feature extraction, AL-GNN performs only a single-step analytic recursive update using Eq.~\eqref{eq:Rt_woodbury} and Eq.~\eqref{eq:Wt_recursive}, without iterative replay optimization or repeated backpropagation during incremental learning. As a result, the practical incremental overhead of AL-GNN is substantially lower than that of optimization-based continual learning frameworks.

% -------------------------- Table: Time Complexity ---------------------------- %

\begin{table}[t]
\centering
\caption{Comparison of additional continual learning overhead among different methods.}
\label{tab:complexity_comparison}
\renewcommand{\arraystretch}{1.3}
% \setlength{\tabcolsep}{2pt}
% \footnotesize
\scalebox{0.86}{
\begin{tabular}{p{1.15cm}p{3cm}p{2.1cm}p{2.2cm}}
\toprule
\textbf{Method} & \textbf{Historical information} & \textbf{Space} & \textbf{Time} \\
\midrule

Bare
& None
& $\mathcal{O}(P_b)$
& $\mathcal{O}(E P_b)$ \\

EWC
& Fisher + parameters
& $\mathcal{O}(P_b)$
& $\mathcal{O}(E P_b)$ \\

LwF
& Previous model
& $\mathcal{O}(P_b)$
& $\mathcal{O}(E P_b)$ \\

MAS
& Importance statistics
& $\mathcal{O}(P_b)$
& $\mathcal{O}(E P_b)$ \\

TWP
& Topology-aware statistics
& $\mathcal{O}(P_b)$
& $\mathcal{O}(E P_b)$ \\

GEM
& Replay nodes
& $\mathcal{O}(Md)$
& $\mathcal{O}(EMP_b)$ \\

ERGNN
& Replay subgraphs
& $\mathcal{O}(Md+E_M)$
& $\mathcal{O}(E(Md+E_M))$ \\

CaT
& Condensed replay graph
& $\mathcal{O}(M_c(d+\bar{k}))$
& $\mathcal{O}(EM_c(d+\bar{k}))$ \\

PI-GNN
& Isolated parameters
& $\mathcal{O}(TP_b)$
& $\mathcal{O}(E P_b)$ \\

CGNN
& Replay graph information
& $\mathcal{O}(M(d+\bar{k}))$
& $\mathcal{O}(EM(d+\bar{k}))$ \\

AL-GNN
& Analytic memory $\mathbf{R}_t$
& $\mathcal{O}(R)$
& $\mathcal{O}(R)$ \\

\bottomrule
\end{tabular}
}
\end{table}

\paragraph{Replay-free analytic memory.}
Unlike replay-based methods that must retain historical graph data (e.g., raw node features, adjacency relations, or condensed subgraphs) to mitigate catastrophic forgetting, AL-GNN stores only the analytic memory matrix $\mathbf{R}_t$, which aggregates second-order feature covariance statistics in the expanded feature space. As summarized in Table~\ref{tab:complexity_comparison}, the memory complexity of AL-GNN is $\mathcal{O}(R)$ with $R=d_{\mathrm{feg}}^2$, independent of the number of historical nodes, replay edges, or incremental sessions. The retained $\mathbf{R}_t$ encodes only aggregated statistical information without preserving explicit node identities or graph structures (see Algorithm~\ref{alg:algnn}), making it difficult to reconstruct original historical data from the stored memory. Consequently, AL-GNN behaves as a fixed-capacity analytic memory mechanism rather than an ever-growing replay storage framework, offering a lightweight and replay-free solution for class-incremental graph learning.

\begin{figure*}[t]
    \centering
    \includegraphics[width=0.75\textwidth]{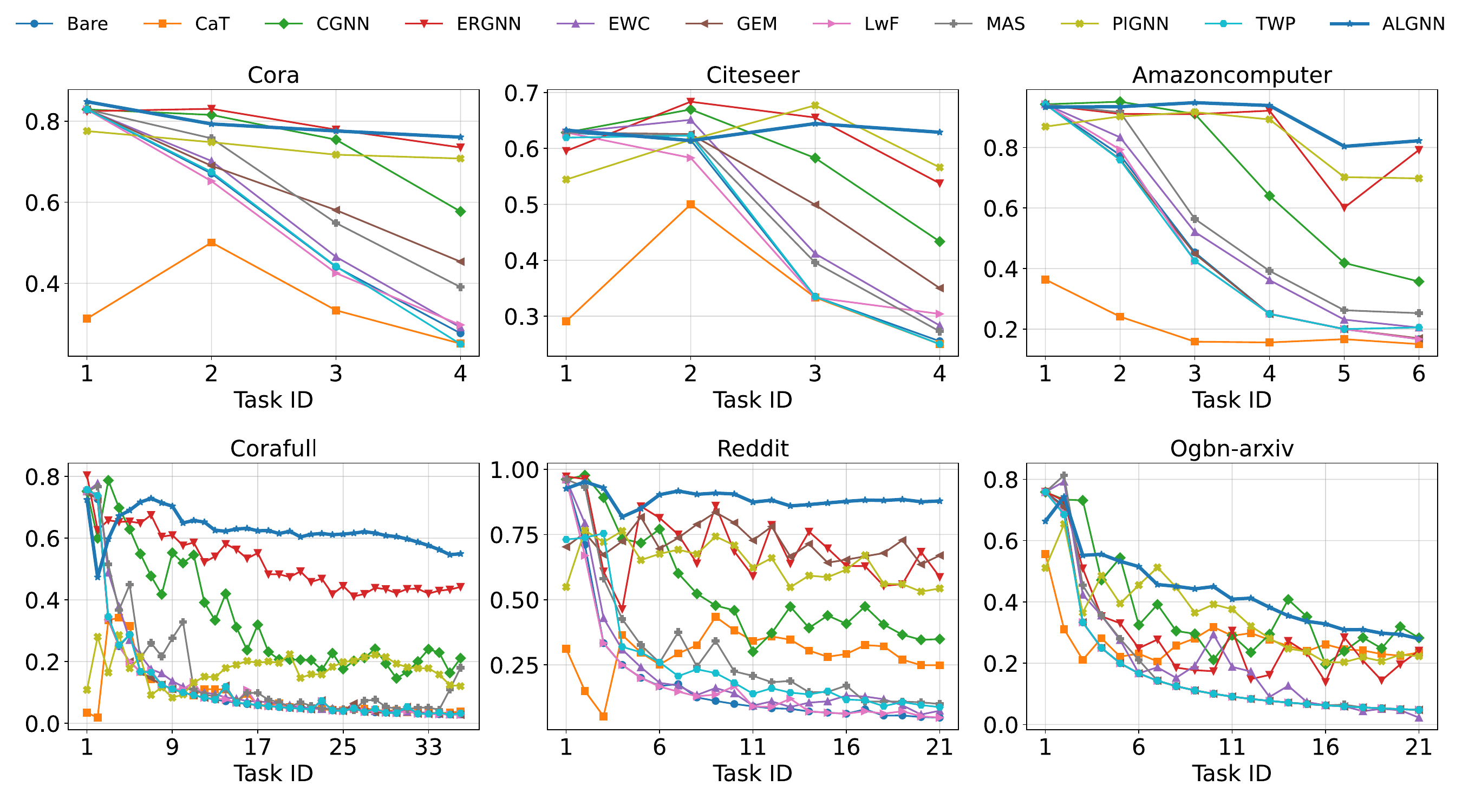}
    \caption{\textbf{Stage-wise performance (AP) over incremental steps across six graph datasets.} The proposed AL-GNN consistently outperforms the competing baselines in final AP across all settings, and maintains stable performance under long incremental horizons (e.g., Corafull, Reddit, Ogbn-arxiv).}
    \label{fig:all_datasets}
\end{figure*}

\section{Experiments}

\subsection{Experimental Setup}

\textbf{Basic Setup}. We evaluate AL-GNN on six benchmark graph datasets: Cora~\cite{2023_AAAI_GREET_Beyond-Smoothing=Unsupervised-Graph-Representation-Learning-with-Edge-Heterophily-Discriminating,2024_SSRN_CANNON_No-Fear-of-Representation-Bias=Graph-Contrastive-Learning-with-Calibration-and-Fusion}, Citeseer~\cite{1998_Digital-Libraries_CiteSeer=An-Automatic-Citation-Indexing-System}, Corafull~\cite{2017_arXiv_Deep-Gaussian-Embedding-of-Graphs=Unsupervised-Inductive-Learning-via-Ranking}, Reddit~\cite{2017_NeurIPS-GraphSAGE_Inductive-Representation-Learning-on-Large-Graphs}, Ogbn-arxiv~\cite{2020_NeurIPS_Open-Graph-Benchmark=Datasets-for-Machine-Learning-on-Graphs}, and Amazon Computer~\cite{2018_arXiv_Pitfalls-of-Graph-Neural-Network-Evaluation}. We compare against Bare model (no strategy), EWC~\cite{2017_NAS_EWC_Overcoming-Catastrophic-Forgetting-in-Neural-Networks}, LwF~\cite{2017_TPAMI_LwF_Learning-without-Forgetting}, MAS~\cite{2018_ECCV_MAS_Memory-Aware-Synapses}, TWP~\cite{2021_AAAI_TWP_Overcoming-Catastrophic-Forgetting-in-Graph-Neural-Networks}, GEM~\cite{2017_NeurIPS_GEM_Gradient-Episodic-Memory-for-Continual-Learning}, ERGNN~\cite{2021_AAAI_ER-GNN_OVercoming-Catastrophic-Forgetting-in-Graph-Neural-Networks-with-Experience-Replay}, CaT~\cite{2023_ICDM_CaT_CaT=Balanced-Continual-Graph-Learning-with-Graph-Condensation}, PIGNN~\cite{2023_SIGIR_PI-GNN_Continual-Learning-on-Dynamic-Graphs-via-Parameter-Isolation}, and CGNN~\cite{2020_CIKM_ContinualGNN_Streaming-Graph-Neural-Networks-via-Continual-Learning}. We use an RTX3090 GPU. For all datasets except Corafull and Amazon Computer, we follow the original splits. For Corafull and Amazon Computer, we use a 40\%/10\%/50\% train/validation/test split.

\textbf{Implementation Details}. We adopt a 2-layer GCN backbone with 256 hidden units. Baseline methods are trained for 50 epochs per stage with Adam, a learning rate of 0.001, dropout 0.5, and weight decay $5 \times 10^{-4}$. For AL-GNN, backpropagation is used only in the base stage. The encoder is then frozen and all later updates are analytic. The feature expansion layer is initialized once and kept fixed. The expansion dimension is set to $2048$ for Cora, Citeseer, and Corafull, $8192$ for Ogbn-arxiv, and $1024$ for Reddit and Amazon Computer. The regularization coefficient $\gamma$ is selected on the validation set.

\textbf{Protocol and Fairness}. All methods are evaluated under the same class-incremental data stream, backbone capacity, and dataset splits unless a method requires a different routine by design. After each incremental step, evaluation is performed on all classes observed so far. All experiments use a fixed random seed of 42, and we report average performance with a 95\% confidence interval over repeated runs.

\textbf{Evaluation Metrics}.
To comprehensively evaluate the performance of AL-GNN and baseline methods, we adopt three metrics:

\begin{itemize}
\item \textbf{Average Performance}~\cite{zhang2022cglb}. Measures the average accuracy across all seen sessions after training, reflecting the model’s ability to retain and integrate knowledge. Higher is better.

\item \textbf{Average Forgetting}~\cite{zhang2022cglb}. Calculates the average performance drop for each session from when it was learned to after the final session, indicating forgetting. Lower is better.

\item \textbf{Training Time}. Records the total time consumed in all phases, including base and incremental training. Measured in seconds, it reflects computational efficiency.
\end{itemize}

Formally, AP and AF are computed as:

\begin{equation}
\text{AP}^{(t)} = \frac{1}{t+1} \sum_{i=0}^{t} \mathrm{M}_{t,i}, \quad
\text{AF}^{(t)} = \frac{1}{t} \sum_{i=0}^{t-1} \left(\mathrm{M}_{i,i} - \mathrm{M}_{t,i}\right),
\end{equation}

where $\mathrm{M}_{t,i}$ denotes the model's accuracy on increment $i$ after training up to increment $t$. In the final summary tables, we report the final-step values $\text{AP}^{(T)}$ and $\text{AF}^{(T)}$.

% ----------------------- Table: overall performance --------------------------- %

\begin{table*}[t]
% \begingroup
% \tiny
% \setlength{\tabcolsep}{1.5pt}
\renewcommand{\arraystretch}{1.6}
\caption{Performance comparison (mean\textsubscript{±std}) of continual learning methods on six datasets using AP, AF, and runtime.}
% \resizebox{\textwidth}{!}{
\centering
\scalebox{0.78}{
% \begin{tabular}{llcccccccccc>{\columncolor{myblue}}c}
\begin{tabular}{llccccccccccc}
\toprule
\textbf{Dataset} & \textbf{Metric} & \textbf{Bare} & \textbf{CaT} & \textbf{CGNN} & \textbf{ERGNN} & \textbf{EWC} & \textbf{GEM} & \textbf{LwF} & \textbf{MAS} & \textbf{PIGNN} & \textbf{TWP} & \textbf{ALGNN} \\
\midrule
\multirow{3}{*}{Cora}
& AP   & 27.67\textsubscript{±3.78} & 25.12\textsubscript{±0.17} & 57.71\textsubscript{±3.11} & 73.50\textsubscript{±1.49} & 29.11\textsubscript{±1.69} & 45.34\textsubscript{±0.30} & 29.73\textsubscript{±6.69} & 39.07\textsubscript{±6.75} & 70.78\textsubscript{±0.31} & 25.03\textsubscript{±1.04} & \textbf{75.86\textsubscript{±1.76}} \\
& AF   & 90.74\textsubscript{±4.83} & 76.93\textsubscript{±7.40} & 44.74\textsubscript{±2.38} & 18.60\textsubscript{±2.15} & 88.60\textsubscript{±2.77} & 65.61\textsubscript{±2.43} & 87.99\textsubscript{±8.72} & 74.59\textsubscript{±3.14} & 28.13\textsubscript{±1.92} & 94.30\textsubscript{±0.21} & \textbf{9.81\textsubscript{±2.06}} \\
& Time & 5.90\textsubscript{±0.4} & 8.21\textsubscript{±0.2} & 5.82\textsubscript{±0.2} & 6.16\textsubscript{±0.3} & 6.93\textsubscript{±0.4} & 19.64\textsubscript{±0.2} & 7.34\textsubscript{±0.8} & 6.52\textsubscript{±0.1} & 7.03\textsubscript{±0.4} & 8.76\textsubscript{±0.3} & \textbf{1.08\textsubscript{±0.11}} \\
\midrule
\multirow{3}{*}{Citeseer}
& AP   & 25.52\textsubscript{±0.73} & 25.02\textsubscript{±0.41} & 43.35\textsubscript{±1.56} & 53.76\textsubscript{±2.42} & 28.32\textsubscript{±0.84} & 35.02\textsubscript{±4.59} & 30.40\textsubscript{±7.64} & 27.27\textsubscript{±0.44} & 56.60\textsubscript{±0.75} & 25.03\textsubscript{±0.13} & \textbf{64.15\textsubscript{±4.31}} \\
& AF   & 86.49\textsubscript{±1.12} & 76.36\textsubscript{±5.46} & 52.56\textsubscript{±3.47} & 36.50\textsubscript{±3.18} & 81.17\textsubscript{±1.79} & 72.38\textsubscript{±7.49} & 80.20\textsubscript{±1.22} & 76.97\textsubscript{±1.14} & 32.36\textsubscript{±4.86} & 87.09\textsubscript{±0.06} & \textbf{20.51\textsubscript{±2.13}} \\
& Time & 8.30\textsubscript{±0.4} & 13.97\textsubscript{±0.2} & 11.03\textsubscript{±0.6} & 10.18\textsubscript{±0.4} & 10.10\textsubscript{±0.4} & 35.66\textsubscript{±7.4} & 13.08\textsubscript{±0.5} & 14.54\textsubscript{±0.5} & 15.47\textsubscript{±0.3} & 20.90\textsubscript{±5.4} & \textbf{0.53\textsubscript{±0.61}} \\
\midrule
\multirow{3}{*}{Corafull}
& AP   & 3.70\textsubscript{±0.03} & 3.88\textsubscript{±0.09} & 21.09\textsubscript{±0.87} & 42.83\textsubscript{±0.01} & 2.92\textsubscript{±0.02} & 2.78\textsubscript{±1.78} & 2.78\textsubscript{±0.01} & 18.00\textsubscript{±1.09} & 12.03\textsubscript{±0.66} & 3.24\textsubscript{±0.65} & \textbf{55.21\textsubscript{±1.15}} \\
& AF   & 98.34\textsubscript{±0.02} & 86.60\textsubscript{±4.49} & 72.52\textsubscript{±0.49} & 43.51\textsubscript{±0.03} & 99.14\textsubscript{±0.04} & 99.24\textsubscript{±4.83} & 99.27\textsubscript{±0.05} & 79.96\textsubscript{±1.18} & 21.30\textsubscript{±8.29} & 98.83\textsubscript{±0.67} & \textbf{4.62\textsubscript{±0.05}} \\
& Time & 309.34\textsubscript{±6.8} & 324.00\textsubscript{±1.8} & 348.36\textsubscript{±12.6} & 318.17\textsubscript{±5.7} & 392.61\textsubscript{±8.8} & 1969.90\textsubscript{±35.2} & 431.26\textsubscript{±0.5} & 315.33\textsubscript{±0.1} & 423.36\textsubscript{±4.4} & 383.17\textsubscript{±3.6} & \textbf{7.33\textsubscript{±0.11}} \\
\midrule
\multirow{3}{*}{Reddit}
& AP   & 4.76\textsubscript{±0.72} & 24.85\textsubscript{±4.00} & 34.87\textsubscript{±1.07} & 58.68\textsubscript{±0.90} & 7.42\textsubscript{±0.64} & 66.96\textsubscript{±0.25} & 4.76\textsubscript{±0.15} & 10.10\textsubscript{±0.36} & 54.31\textsubscript{±1.65} & 8.79\textsubscript{±0.45} & \textbf{87.70\textsubscript{±0.30}} \\
& AF   & 99.81\textsubscript{±0.42} & 32.37\textsubscript{±6.69} & 66.98\textsubscript{±1.12} & 42.18\textsubscript{±0.95} & 96.99\textsubscript{±0.67} & 16.95\textsubscript{±0.25} & 99.80\textsubscript{±0.63} & 94.06\textsubscript{±0.53} & 37.91\textsubscript{±0.13} & 93.89\textsubscript{±0.85} & \textbf{1.37\textsubscript{±1.15}} \\
& Time & 1212.69\textsubscript{±6.9} & 1157.81\textsubscript{±9.9} & 1267.37\textsubscript{±36.0} & 1188.48\textsubscript{±11.0} & 1207.84\textsubscript{±40.5} & 4580.42\textsubscript{±95.8} & 1661.33\textsubscript{±2.5} & 1164.92\textsubscript{±15.2} & 1231.50\textsubscript{±8.2} & 862.65\textsubscript{±0.2} & \textbf{28.45\textsubscript{±0.94}} \\
\midrule
\multirow{3}{*}{\shortstack[l]{Ogbn \\ Arxiv}}
& AP   & 4.76\textsubscript{±0.13} & 24.78\textsubscript{±2.11} & 28.05\textsubscript{±0.41} & 23.19\textsubscript{±0.12} & 2.23\textsubscript{±1.28} & 4.73\textsubscript{±0.04} & 4.77\textsubscript{±0.31} & 4.76\textsubscript{±0.29} & 22.27\textsubscript{±1.08} & 4.73\textsubscript{±0.51} & \textbf{28.08\textsubscript{±0.22}} \\
& AF   & 98.76\textsubscript{±0.21} & 12.09\textsubscript{±1.68} & 58.73\textsubscript{±0.35} & 76.01\textsubscript{±0.23} & 83.66\textsubscript{±0.12} & 98.76\textsubscript{±0.64} & 98.74\textsubscript{±0.63} & 98.15\textsubscript{±0.07} & 57.62\textsubscript{±2.15} & 98.78\textsubscript{±0.53} & \textbf{6.37\textsubscript{±0.21}} \\
& Time & 89.85\textsubscript{±24.5} & 97.21\textsubscript{±0.0} & 78.07\textsubscript{±1.1} & 74.18\textsubscript{±0.7} & 95.78\textsubscript{±0.8} & 497.57\textsubscript{±4.2} & 97.21\textsubscript{±0.0} & 74.73\textsubscript{±0.0} & 117.85\textsubscript{±2.0} & 105.23\textsubscript{±1.4} & \textbf{27.76\textsubscript{±0.32}} \\
\midrule
\multirow{3}{*}{\shortstack[l]{Amazon\\Computer}}
& AP   & 16.67\textsubscript{±0.41} & 15.03\textsubscript{±3.61} & 35.71\textsubscript{±0.09} & 79.19\textsubscript{±0.08} & 20.50\textsubscript{±0.18} & 16.94\textsubscript{±0.38} & 16.67\textsubscript{±0.32} & 25.26\textsubscript{±0.94} & 69.80\textsubscript{±1.49} & 20.57\textsubscript{±4.36} & \textbf{80.15\textsubscript{±4.60}} \\
& AF   & 98.85\textsubscript{±0.08} & 9.26\textsubscript{±2.86} & 73.29\textsubscript{±0.44} & 22.73\textsubscript{±0.28} & 94.25\textsubscript{±0.13} & 98.53\textsubscript{±0.54} & 98.86\textsubscript{±0.08} & 88.20\textsubscript{±1.27} & 32.01\textsubscript{±3.06} & 94.17\textsubscript{±5.31} & \textbf{13.78\textsubscript{±6.61}} \\
& Time & 6.65\textsubscript{±0.5} & 10.64\textsubscript{±0.1} & 7.57\textsubscript{±0.1} & 6.87\textsubscript{±0.3} & 8.38\textsubscript{±0.1} & 21.24\textsubscript{±0.3} & 8.55\textsubscript{±0.4} & 7.33\textsubscript{±0.2} & 8.40\textsubscript{±0.3} & 9.88\textsubscript{±0.6} & \textbf{0.39\textsubscript{±0.03}} \\
\bottomrule

\end{tabular}
}
\label{tab:final_alg_results}
% \endgroup
\end{table*}

\begin{figure}
    \centering
    \includegraphics[width=0.45\textwidth]{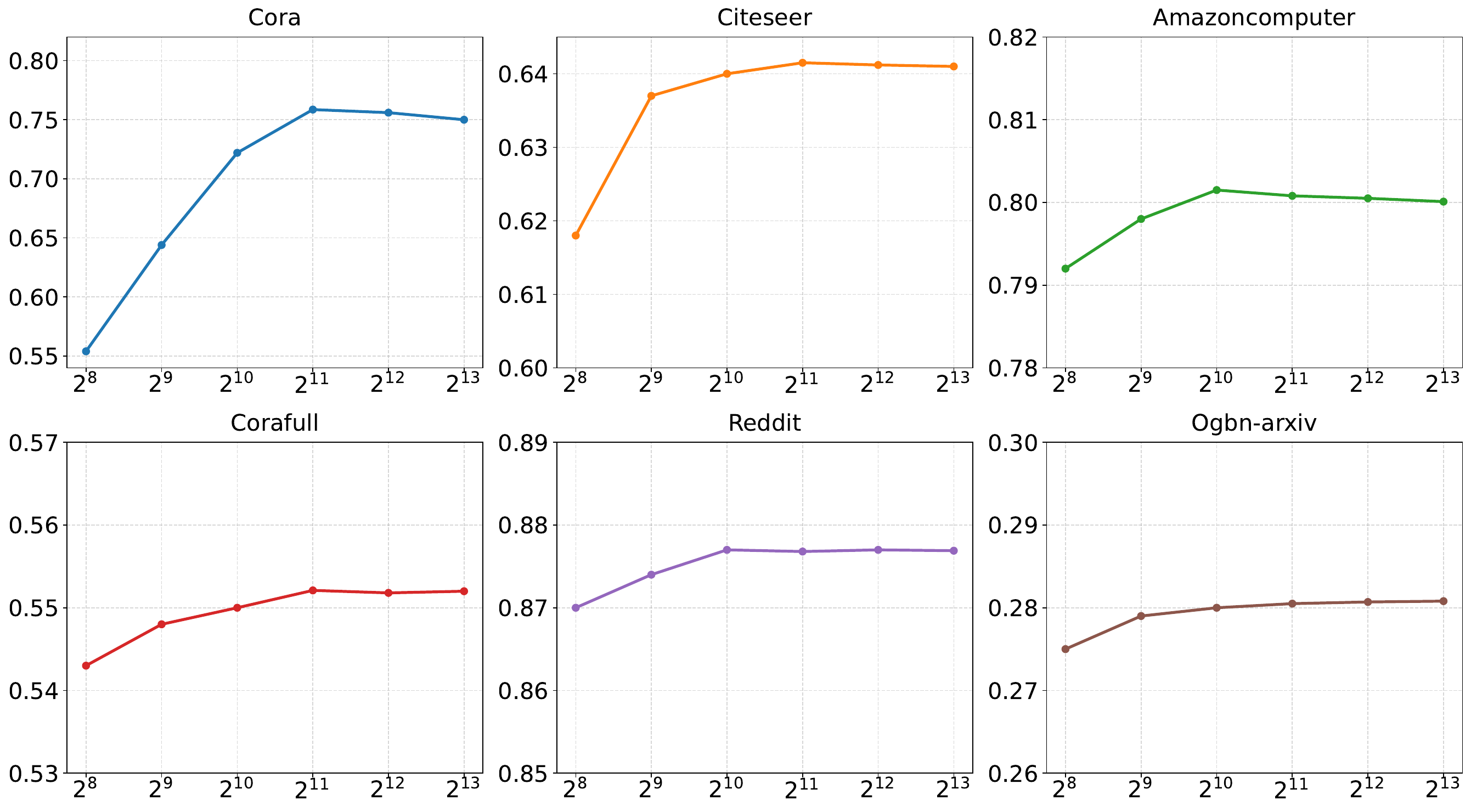}
    \caption{The impact of the feature expansion dimension on incremental learning performance.}
    \label{fig:feg_sensitivity}
\end{figure}

\subsection{Main Results}

Table~\ref{tab:final_alg_results} reports the overall comparison between AL-GNN and ten representative continual graph learning baselines on six benchmark datasets, using average performance (AP), average forgetting (AF), and total training time as evaluation criteria. Overall, AL-GNN delivers the strongest and most consistent performance across datasets. It achieves the best AP on all six benchmarks, including Cora (75.86\%), Citeseer (64.15\%), Corafull (55.21\%), Reddit (87.70\%), Ogbn-arxiv (28.08\%), and Amazon Computer (80.15\%). These results indicate that the proposed analytic update mechanism generalizes well across both small citation networks and larger real-world graphs.

A similar advantage is observed in knowledge retention. AL-GNN attains the lowest or near-lowest AF on most datasets, with particularly strong results on Cora (9.81\%), Citeseer (20.51\%), Corafull (4.62\%), Reddit (1.37\%), and Ogbn-arxiv (6.37\%). On these benchmarks, most competing methods suffer substantial degradation after sequential updates, whereas AL-GNN maintains much stronger retention of previously acquired knowledge. Although AL-GNN is not the best AF performer on every dataset---for example, CaT obtains a lower AF on Amazon Computer---its overall retention remains highly competitive while simultaneously providing the best AP and runtime. This trade-off is important in practice, since a method with low forgetting but clearly inferior final accuracy is less desirable in class-incremental deployment.

\begin{figure}
    \centering
    \includegraphics[width=0.45\textwidth]{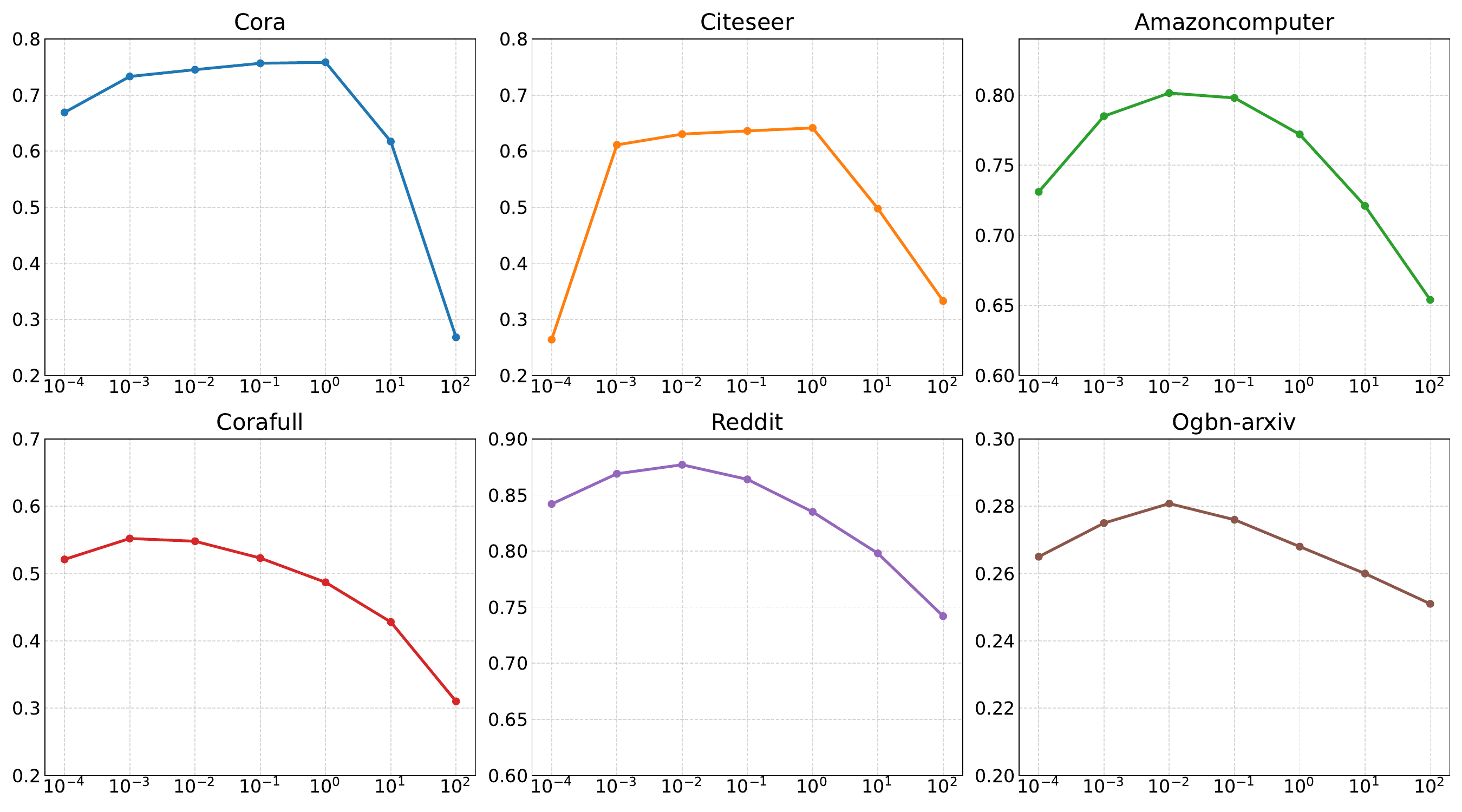}
    \caption{The impact of $\gamma$ on incremental learning performance.}
    \label{fig:gamma_sensitivity}
\end{figure}

In addition to accuracy and retention, AL-GNN is markedly more efficient than all competing methods. Its total training time is the lowest on every benchmark, e.g., 1.08s on Cora, 0.53s on Citeseer, 7.33s on Corafull, 28.45s on Reddit, 27.76s on Ogbn-arxiv, and 0.39s on Amazon Computer. The efficiency gain is particularly evident on datasets with long incremental horizons or expensive repeated optimization, where replay-based and regularization-based methods incur substantial stage-wise training overhead. For instance, on Reddit, GEM requires 4580.42s, whereas AL-GNN completes training in 28.45s. This large reduction in runtime directly reflects the advantage of replacing repeated backpropagation-based optimization with one-pass analytic recursion during incremental stages.

\begin{table}[t]
\centering
\small
\renewcommand{\arraystretch}{1.4}
\caption{Final AP (\%) of all methods under different $k$ settings and datasets.}
\scalebox{0.68}{
\begin{tabular}{lccc|ccc|ccc}
\toprule
\multirow{2}{*}{Method} & \multicolumn{3}{c|}{Corafull} & \multicolumn{3}{c|}{Ogbn-arxiv} & \multicolumn{3}{c}{Reddit} \\
                        & $k=1$ & $k=5$ & $k=10$ & $k=1$ & $k=5$ & $k=10$ & $k=1$ & $k=5$ & $k=10$ \\
\midrule
Bare   & 3.70 & 17.08 & 30.92 & 4.76 & 21.42 & 28.44 & 4.76 & 23.41 & 34.79 \\
CaT    & 3.88 & 10.29 & 17.30 & 24.78 & 18.29 & 39.36 & 24.85 & 9.12 & 22.82 \\
CGNN   & 21.09 & 39.67 & 46.35 & 28.05 & 48.87 & \textbf{54.47} & 34.87 & 42.61 & 66.54 \\
ERGNN  & 44.19 & 28.53 & 41.50 & 24.12 & 27.20 & 29.42 & 58.68 & 29.05 & 35.09 \\
EWC    & 2.92 & 26.60 & 36.19 & 2.23 & 31.66 & 31.18 & 7.42 & 29.70 & 35.64 \\
GEM    & 2.78 & 17.33 & 30.58 & 4.73 & 33.78 & 28.29 & 66.96 & 33.82 & 42.03 \\
LwF    & 2.78 & 15.50 & 30.21 & 4.77 & 19.99 & 28.91 & 4.76 & 25.45 & 34.83 \\
MAS    & 18.00 & 49.63 & \textbf{55.86} & 4.76 & 35.26 & 39.85 & 10.10 & 62.93 & 73.63 \\
PIGNN  & 12.03 & 45.42 & 51.90 & 22.27 & \textbf{53.90} & 54.38 & 54.31 & 74.54 & 79.71 \\
TWP    & 3.24 & 17.04 & 30.96 & 4.73 & 21.27 & 27.94 & 8.79 & 22.38 & 35.11 \\
AL-GNN  & \textbf{55.21} & \textbf{54.70} & 52.80 & \textbf{28.08} & 43.05 & 43.89 & \textbf{87.70} & \textbf{87.27} & \textbf{89.40} \\
\bottomrule
\end{tabular}
}
\label{tab:final_ap_transposed}
\end{table}

Among the baselines, replay-based methods such as ERGNN, GEM, and PIGNN often remain competitive on certain datasets, suggesting that access to historical samples can still be beneficial for alleviating forgetting. However, these methods generally incur substantially higher computational cost and exhibit less consistent behavior across benchmarks. Regularization-based approaches, including EWC, LwF, MAS, and TWP, are typically less competitive in both AP and AF. In contrast, AL-GNN provides a more favorable balance among predictive performance, retention, and efficiency without storing historical graph data.

The dataset-wise results also clarify where the proposed analytic formulation is most advantageous. On citation-style benchmarks such as Cora, Citeseer, and Corafull, AL-GNN combines strong retention with very low runtime, indicating that a stable base-stage representation can support effective analytic adaptation over long label sequences. On Reddit and Ogbn-arxiv, where repeated training is more expensive, the runtime advantage becomes more pronounced while AP remains competitive or best overall. Amazon Computer is more challenging from the forgetting perspective, as certain replay-based methods retain a slight advantage in AF, but AL-GNN still achieves the best final AP and the lowest training time.

\subsection{Stability Under Different Incremental Horizons}

While the results in Table~\ref{tab:final_alg_results} establish the overall advantage of AL-GNN, a more stringent question in continual graph learning is whether such an advantage can be maintained under long and highly fragmented incremental streams. This issue is particularly important in class-incremental settings, where the number of update stages is controlled by the number of newly introduced classes per session, denoted by $k$. Smaller $k$ leads to a longer horizon with more frequent session updates, which amplifies representation drift and cumulative forgetting, whereas larger $k$ shortens the sequence but requires the model to absorb more new classes at each session. Therefore, varying $k$ provides a direct test of whether a continual learner remains stable under different sequential regimes.

\paragraph{Stage-wise stability across datasets.}
Fig.~\ref{fig:all_datasets} first presents a stage-wise view of the incremental process on six benchmarks. Across datasets, AL-GNN maintains smoother performance trajectories than competing methods and exhibits substantially less degradation as the number of seen increments increases. This trend is particularly evident on long-horizon benchmarks such as Corafull, Reddit, and Ogbn-arxiv, where many baselines become progressively unstable or show marked drops in later stages. In contrast, AL-GNN remains comparatively stable throughout the sequence, suggesting that its advantage is not limited to final performance alone but extends to the entire continual learning process.

\paragraph{Effect of increment size.}
To more systematically evaluate this property, we further consider three increment settings, $k \in \{1,5,10\}$, on Corafull, Ogbn-arxiv, and Reddit. The final AP values are summarized in Table~\ref{tab:final_ap_transposed}. A consistent observation is that reducing $k$ makes the learning problem considerably more difficult for most baselines. Under $k=1$, where classes arrive one by one and the number of incremental steps becomes the largest, many methods exhibit stronger instability and more pronounced degradation. As $k$ increases, this effect is generally alleviated because fewer incremental steps reduce the accumulation of forgetting. In contrast, AL-GNN remains substantially less sensitive to the choice of increment size across all three settings, including the most challenging $k=1$ case.

The final AP results in Table~\ref{tab:final_ap_transposed} further support this conclusion. On Corafull, AL-GNN achieves 55.21\%, 54.70\%, and 52.80\% under $k=1$, $k=5$, and $k=10$, respectively, indicating only limited sensitivity to increment granularity. On Reddit, AL-GNN consistently remains the strongest method across all three settings, reaching 87.70\%, 87.27\%, and 89.40\%. Even on Ogbn-arxiv, where certain replay-based methods benefit more from coarser increments, AL-GNN remains highly competitive and exhibits more stable behavior than most regularization-based baselines. These results suggest that the proposed analytic update mechanism does not rely on a specific session partition to remain effective.

\begin{figure*}[t]
    \centering
    \includegraphics[width=0.85\textwidth]{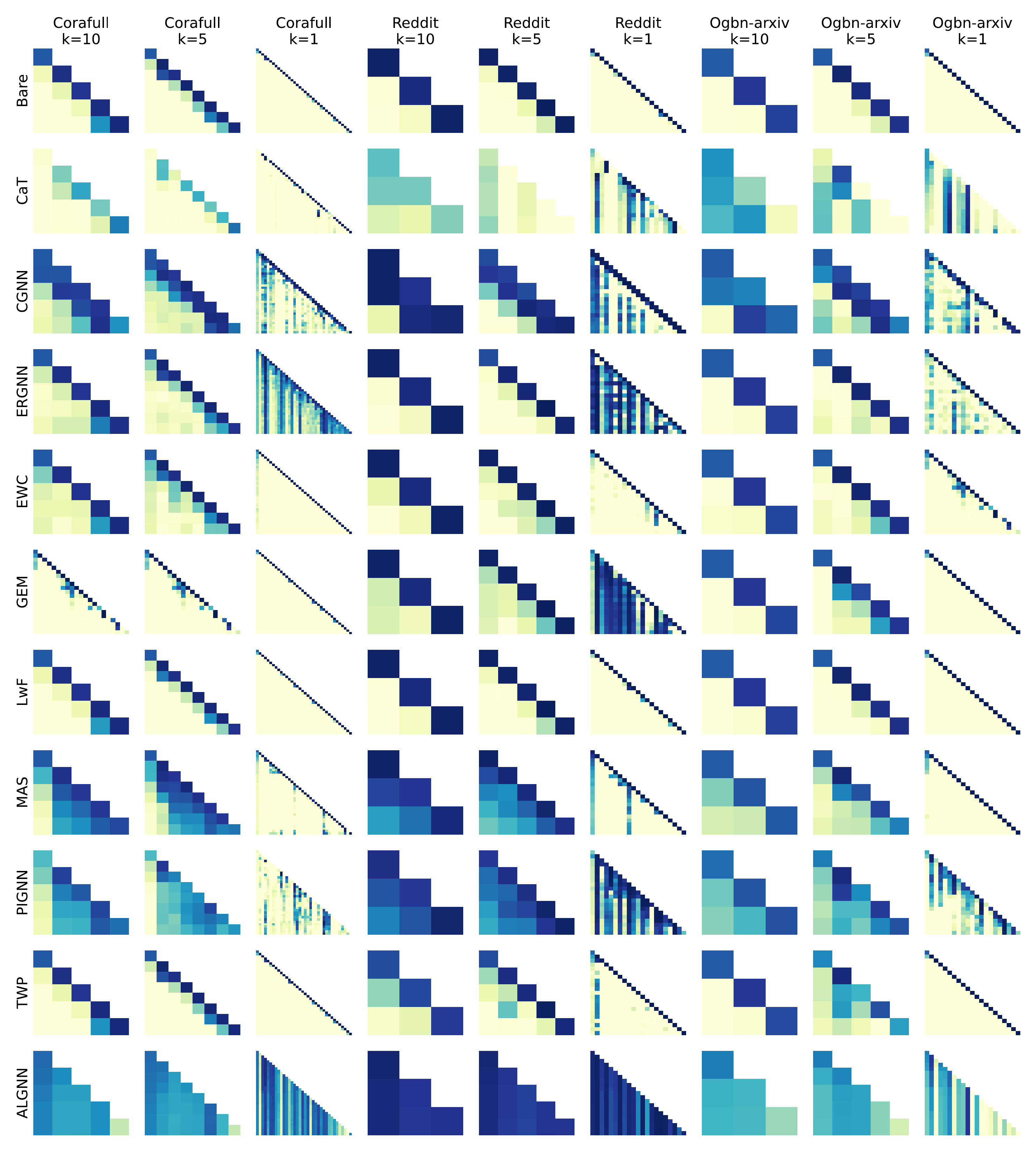}
    \caption{
    Visualization of the lower-triangular AP matrices across different continual learning algorithms.}
    \label{fig:exp_test_all_algos_k_heatmap}
\end{figure*}

\paragraph{Forgetting propagation analysis.}
Fig.~\ref{fig:exp_test_all_algos_k_heatmap} provides additional insight into how forgetting propagates throughout the incremental sequence. For most baselines, the lower-triangular accuracy patterns become increasingly non-uniform as the number of stages grows, especially under smaller $k$. This fading away from the diagonal indicates that the models gradually lose performance on earlier classes after repeated updates. By contrast, AL-GNN preserves a much more uniform lower-triangular structure across datasets and increment settings, indicating that previously learned classes remain comparatively stable even after long update sequences. Importantly, this observation complements the final AP analysis: AL-GNN not only performs well at the end of training, but also maintains more stable retention throughout the entire continual learning process.

Overall, the evidence from stage-wise trajectories, final AP comparisons, and lower-triangular retention maps consistently shows that the main advantage of AL-GNN lies not merely in strong final accuracy, but in its robustness under different incremental horizons.
In practical continual graph learning scenarios, where new classes may arrive either individually or in small batches, such robustness is particularly important. The results therefore suggest that AL-GNN is better suited to long-horizon class-incremental learning than existing replay-based or regularization-based alternatives.

\subsection{Ablation and Resource Analysis}

We further analyze the main design factors of AL-GNN and its resource requirements, focusing on the feature expansion dimension, the regularization parameter $\gamma$, and the memory cost of maintaining the recursive analytic statistics.

\paragraph{Effect of feature expansion dimension.}
Fig.~\ref{fig:feg_sensitivity} analyzes the influence of the feature expansion dimension on incremental performance across six datasets. A consistent observation is that moderate feature expansion is beneficial, but the gain saturates once the expanded representation becomes sufficiently expressive. On Cora, Citeseer, and Corafull, the average performance improves steadily as the expansion dimension increases and then plateaus around 2048--4096. Reddit and Amazon Computer saturate earlier, around 1024, indicating that a lower-dimensional expansion is already sufficient to capture most of the discriminative information in these datasets. By contrast, Ogbn-arxiv is relatively insensitive to the expansion dimension, which is likely due to its richer node attributes and larger-scale graph structure. Overall, these results suggest that feature expansion is important for improving the capacity of the analytic classifier, but excessively large expansion brings diminishing returns.

\paragraph{Effect of the regularization parameter.}
Fig.~\ref{fig:gamma_sensitivity} evaluates the effect of the regularization parameter $\gamma$. Across datasets, the performance exhibits a broadly unimodal trend: it improves as $\gamma$ increases from very small values, reaches a stable optimal region, and then degrades under excessive regularization. This behavior indicates that $\gamma$ plays an important role in balancing adaptability to newly introduced classes against stability of the recursive least-squares update. The optimal range is relatively narrow on citation datasets such as Cora and Citeseer, whereas Reddit and Amazon Computer show a wider plateau and are therefore more tolerant to the choice of $\gamma$. Ogbn-arxiv again appears comparatively insensitive.

\paragraph{Memory and storage cost.}
The retained-state analysis is discussed in Methodology. Here we report only the instantiated storage of the dominant analytic statistic for the feature dimensions used in our experiments. For the configurations used in our experiments, Cora, Citeseer, and Corafull with $d_{\text{feg}}=2048$ require approximately 16 MB in float32, while Reddit and Amazon Computer with $d_{\text{feg}}=1024$ require approximately 4 MB. Ogbn-arxiv with $d_{\text{feg}}=8192$ requires about 268 MB. Although this larger matrix increases the memory requirement on Ogbn-arxiv, the retained state remains independent of stream length and historical sample count.

Overall, the ablation results indicate that AL-GNN is not overly sensitive to its main design choices as long as the feature expansion dimension and regularization strength are chosen within reasonable ranges.

\section{Discussion}
AL-GNN is motivated by the observation that, given a sufficiently informative graph encoder, continual classification can be reduced to recursive linear estimation in the representation space. Rather than treating forgetting as a data availability issue---as in replay-based methods---AL-GNN frames it as an information aggregation problem, maintaining compact second-order statistics that losslessly summarize all previously observed feature--label correlations under the fixed-feature assumption. A key theoretical consequence is that the recursive analytic update preserves exact equivalence to the joint least-squares solution over all seen sessions, providing a principled explanation for the strong retention behavior observed in our experiments.

This theoretical guarantee, however, depends critically on the frozen-encoder assumption. When later sessions introduce substantially novel graph patterns not captured by the base-stage encoder, the fixed representation space may lack expressiveness, limiting plasticity. The relatively modest absolute performance on Ogbn-arxiv (28.08\% AP) may partially reflect this limitation, as the large and heterogeneous nature of that dataset places higher demands on representation capacity than what a single base-stage GCN can provide.

Several directions for future work emerge from this analysis. First, hybrid approaches combining analytic classifier updates with controlled, low-rank encoder adaptation could extend the theoretical guarantees to settings with moderate distribution shift. Second, stronger graph encoders---such as graph foundation models or pretrained graph transformers---may provide richer initial representations that further enhance the robustness of replay-free analytic learning. Third, extending the framework to handle graph structure evolution (e.g., new edges or nodes arriving without class labels) would broaden its applicability to more general dynamic graph settings.

\section{Conclusion}

This paper presented \textbf{AL-GNN}, a replay-free analytic continual learning framework for class-incremental graph node classification. By decoupling representation learning from classifier adaptation, AL-GNN trains a graph encoder once in the base stage and performs all subsequent updates through closed-form recursive equations in an expanded feature space. The only state maintained across sessions is a fixed-size analytic memory matrix, eliminating replay buffers, iterative backpropagation, and explicit parameter regularization during incremental learning.

AL-GNN provides three main contributions: a principled replay-free solution via recursive least-squares updates, a theoretical equivalence guarantee to the joint analytic solution under frozen features, and strong empirical results on six benchmarks---achieving the best average performance, competitive forgetting resistance, and substantially lower training time compared with ten representative baselines.

\bibliography{main}
\bibliographystyle{IEEEtran}

%===========================================
\vspace{-3em}
\begin{IEEEbiographynophoto}{Xuling Zhang}
is currently working as a Research Assistant at The Hong Kong University of Science and Technology (Guangzhou). He received his M.Sc. degree from Universiti Sains Malaysia. His research interests include multimodal learning, large language models, graph learning, and continual learning. He has published papers in top-tier venues such as ICML and ICASSP.
\end{IEEEbiographynophoto}
\vspace{-3em}
\begin{IEEEbiographynophoto}{Jindong Li}
is currently pursuing the Ph.D. degree at The Hong Kong University of Science and Technology (Guangzhou). His research interests include multimodal learning, large language models, and graph learning. He has published papers in top-tier venues such as TPAMI, ICLR, ACL, AAAI, IJCAI, ECML PKDD, and Neural Networks.
\end{IEEEbiographynophoto}
\vspace{-3em}
\begin{IEEEbiographynophoto}{Yifei Zhang}
is a Professor at Northwestern Polytechnical University (NPU). Prior to joining NPU, he worked as a Research Scientist at Nanyang Technological University (NTU), Singapore. He received his Ph.D. degree from The Chinese University of Hong Kong (CUHK). His research interests include representation learning, graph learning, and large language models. He has published extensively in leading AI and data science venues, including NeurIPS, CVPR, ICLR, AAAI, IJCAI, and KDD.
\end{IEEEbiographynophoto}
\vspace{-3em}
\begin{IEEEbiographynophoto}{Mingqi Yang}
is an assistant professor at the Department of Electronic Business, South China University of Technology, Guangzhou, China. He received the PhD. degree from Dalian University of Technology and was a Postdoctoral Research Fellow at the National University of Singapore. His research interests include graph machine learning, trustworthy AI, AI for science, and business intelligence. He has published papers in top-tier venues such as ICML, KDD, WWW, AAAI, and TKDE.
\end{IEEEbiographynophoto}
\vspace{-3em}
\begin{IEEEbiographynophoto}{Menglin Yang}
is an Assistant Professor with the Thrust of Artificial Intelligence, The Hong Kong University of Science and Technology (Guangzhou). He received the Ph.D. degree from The Chinese University of Hong Kong and was a Postdoctoral Researcher at Yale University. His research interests include hyperbolic representation learning, multimodal learning, and information retrieval. He has published papers in top-tier venues, including NeurIPS, ICML, CVPR, KDD, WWW, SIGIR, TKDE, and ICDE.
\end{IEEEbiographynophoto}
\vfill
%============================================
\onecolumn

\appendix

\section{Technical Appendices and Supplementary Material}
\subsection{Regularized Feature Autocorrelation Matrix}

We aim to prove the recursive formulation of the regularized feature autocorrelation matrix \( \mathbf{R}_t \), which plays a key role in the analytic continual learning update. The matrix is defined based on all expanded feature observations seen up to session \( t \) as follows:

\begin{equation}
\mathbf{R}_t = \left( \sum_{s=0}^{t} \mathbf{Z}_s^\top \mathbf{Z}_s + \gamma \mathbf{I} \right)^{-1}.
\end{equation}

This inverse-regularized Gram matrix captures second-order statistics of the expanded features across all seen sessions. We now show that \( \mathbf{R}_t \) admits a recursive formulation:

\begin{equation}
\mathbf{R}_t = \Bigl( \mathbf{R}_{t-1}^{-1} + \mathbf{Z}_t^\top \mathbf{Z}_t \Bigr)^{-1}
\label{eq:a}
\end{equation}

This recursive formulation is crucial in continual learning settings where session data arrives incrementally. It allows us to update the regularized autocorrelation matrix efficiently without recomputing it from scratch at each new session.

\vspace{2mm}
\noindent
To prove Eq.~\eqref{eq:a}, we start by expanding the definition of \( \mathbf{R}_t \) using the summation split between previous sessions and the current session \( t \):

\begin{align}
\mathbf{R}_t 
&= \Biggl( \sum_{s=0}^{t} \mathbf{Z}_s^\top \mathbf{Z}_s + \gamma \mathbf{I} \Biggr)^{-1} \notag \\
&= \Biggl( \Bigl[ \sum_{s=0}^{t-1} \mathbf{Z}_s^\top \mathbf{Z}_s + \gamma \mathbf{I} \Bigr] + \mathbf{Z}_t^\top \mathbf{Z}_t \Biggr)^{-1}
\label{eq:b}
\end{align}

Now we denote the inverse of the accumulated feature covariance matrix up to session \( t-1 \) as:

\begin{equation}
\mathbf{R}_{t-1} = \Biggl( \sum_{s=0}^{t-1} \mathbf{Z}_s^\top \mathbf{Z}_s + \gamma \mathbf{I} \Biggr)^{-1}
\label{eq:c}
\end{equation}

Equivalently, we can write its inverse as:

\begin{equation}
\mathbf{R}_{t-1}^{-1} = \sum_{s=0}^{t-1} \mathbf{Z}_s^\top \mathbf{Z}_s + \gamma \mathbf{I} .
\end{equation}

To simplify notation, let us define:

\begin{equation}
\mathbf{A} = \mathbf{R}_{t-1}^{-1}, \quad
\mathbf{B} = \mathbf{Z}_t^\top \mathbf{Z}_t .
\end{equation}

Substituting these into Eq.~\eqref{eq:b}, we obtain:

\begin{equation}
\mathbf{R}_t = \left( \mathbf{A} + \mathbf{B} \right)^{-1} = \left( \mathbf{R}_{t-1}^{-1} + \mathbf{Z}_t^\top \mathbf{Z}_t \right)^{-1}.
\end{equation}

This completes the proof of the recursive relation given in Eq.~\eqref{eq:a}.

\vspace{1mm}
\noindent
This formulation offers two key advantages:
\begin{itemize}
    \item \textbf{Computational efficiency:} the update from \( \mathbf{R}_{t-1} \) to \( \mathbf{R}_t \) requires only a matrix addition and one inverse operation.
    \item \textbf{Memory efficiency:} past data does not need to be retained; only \( \mathbf{R}_{t-1} \) needs to be stored and updated.
\end{itemize}

Such recursive regularized covariance updates are widely used in analytic continual learning (AL) and recursive least squares (RLS) methods to enable fast, stable parameter updates without backpropagation.

\subsection{Woodbury Matrix Expansion of $\mathbf{R}_t$}

To reduce the cost of inverting large matrices when updating the regularized feature autocorrelation matrix \( \mathbf{R}_t \), we leverage the Woodbury matrix identity, which is defined as:

\begin{equation}
(A + U C V)^{-1} = A^{-1} - A^{-1} U \left( C^{-1} + V A^{-1} U \right)^{-1} V A^{-1}
\label{eq:d}
\end{equation}

In our setting, we define the terms as follows:

\begin{equation}
\mathbf{A} = \mathbf{R}_{t-1}^{-1}, \quad
U = \mathbf{Z}_t^\top, \quad
V = \mathbf{Z}_t, \quad
C = \mathbf{I} .
\end{equation}

Recall from Eq.~\eqref{eq:a} that the recursive definition of \( \mathbf{R}_t \) is:

\begin{equation}
\mathbf{R}_t = \left( \mathbf{R}_{t-1}^{-1} + \mathbf{Z}_t^\top \mathbf{Z}_t \right)^{-1} .
\end{equation}

This is in the exact form of \( (A + U C V)^{-1} \). Thus, applying Eq.~\eqref{eq:d}, we obtain:

\begin{equation}
\mathbf{R}_t =
\mathbf{R}_{t-1} - \mathbf{R}_{t-1} \mathbf{Z}_t^\top
\Bigl( \mathbf{I} + \mathbf{Z}_t \mathbf{R}_{t-1} \mathbf{Z}_t^\top \Bigr)^{-1}
\mathbf{Z}_t \mathbf{R}_{t-1}
\label{eq:e}
\end{equation}

\vspace{2mm}
\noindent
This expression allows \( \mathbf{R}_t \) to be updated recursively without directly computing any matrix inversion from scratch. Instead, it reduces the inversion operation to a much smaller matrix of size \( n_t \times n_t \), where \( n_t \) is the number of labeled nodes in session \( t \), and \( n_t \ll d_{\mathrm{feg}} \) (the expanded feature dimension).

\vspace{2mm}
\noindent
To make this more concrete, we define:

\begin{equation}
\mathbf{P}_t = \left( \mathbf{I} + \mathbf{Z}_t \mathbf{R}_{t-1} \mathbf{Z}_t^\top \right)^{-1} .
\end{equation}

Then Eq.~\eqref{eq:e} becomes:

\begin{equation}
\mathbf{R}_t = \mathbf{R}_{t-1} - \mathbf{R}_{t-1} \mathbf{Z}_t^\top \mathbf{P}_t \mathbf{Z}_t \mathbf{R}_{t-1} .
\end{equation}

This update requires only a rank update on \( \mathbf{R}_{t-1} \), making the computation extremely efficient in online or session-incremental settings. It is especially well-suited for analytic continual learning frameworks, where closed-form updates are preferred over costly gradient-based backpropagation.

\vspace{2mm}
\noindent
Additionally, when \( \mathbf{Z}_t \) has full row rank and few samples, this update is numerically more stable and memory efficient than directly inverting the large matrix in Eq.~\eqref{eq:a}.

\noindent
This makes the Woodbury-expanded form not only theoretically elegant but also practically valuable in large-scale incremental learning scenarios.

\subsection{Classifier Weights}

We aim to prove the recursive formulation of the classifier weight matrix in AL-GNN. Specifically, we show that the classifier at session \( t \) can be constructed by recursively updating the classifier obtained at session \( t{-}1 \), using the current session's features and labels.

Let us define the accumulated feature--label correlation matrix up to session \( t{-}1 \) as:

\begin{equation}
\mathbf{Q}_{t-1} = 
\left[ \mathbf{Z}_0^\top \mathbf{Y}_0^{\mathrm{train}} 
\ \ 
\mathbf{Z}_1^\top \mathbf{Y}_1^{\mathrm{train}} 
\ \ 
\cdots 
\ \
\mathbf{Z}_{t-2}^\top \mathbf{Y}_{t-2}^{\mathrm{train}} 
\ \
\mathbf{Z}_{t-1}^\top \mathbf{Y}_{t-1}^{\mathrm{train}} \right] .
\end{equation}

According to the closed-form solution for analytic weight matrices and the recursive form of the regularized autocorrelation matrix, we can write:

\begin{equation}
\begin{aligned}
\hat{\mathbf{W}}_{\mathrm{cls}}^{(t)} 
&=
\mathbf{R}_t \left[
\mathbf{Q}_{t-1} \quad \mathbf{Z}_t^\top \mathbf{Y}_t^{\mathrm{train}}
\right]\\
&= \left[
\mathbf{R}_t \mathbf{Q}_{t-1} \quad \mathbf{R}_t \mathbf{Z}_t^\top \mathbf{Y}_t^{\mathrm{train}}
\right] .
\end{aligned}
\label{eq:closed-from-solution}
\end{equation}

The core of this recursive construction lies in evaluating \( \mathbf{R}_t \mathbf{Q}_{t-1} \) efficiently using the previously computed state. By applying the Woodbury identity (see Eq.~\eqref{eq:e}), we expand the term as:

\begin{equation}
\mathbf{R}_t \mathbf{Q}_{t-1} =
\Bigl[ \mathbf{R}_{t-1} - \mathbf{R}_{t-1} \mathbf{Z}_t^\top
\Bigl( \mathbf{I} + \mathbf{Z}_t \mathbf{R}_{t-1} \mathbf{Z}_t^\top \Bigr)^{-1}
\mathbf{Z}_t \mathbf{R}_{t-1} \Bigr] \mathbf{Q}_{t-1}
\label{eq:f}
\end{equation}

To further simplify, we use the identity:

\begin{align}
\mathbf{I} 
&= \left( \mathbf{I} + \mathbf{Z}_t \mathbf{R}_{t-1} \mathbf{Z}_t^\top \right)^{-1} 
   \left( \mathbf{I} + \mathbf{Z}_t \mathbf{R}_{t-1} \mathbf{Z}_t^\top \right) \notag \\
&= \left( \mathbf{I} + \mathbf{Z}_t \mathbf{R}_{t-1} \mathbf{Z}_t^\top \right)^{-1}
   + \left( \mathbf{I} + \mathbf{Z}_t \mathbf{R}_{t-1} \mathbf{Z}_t^\top \right)^{-1}
     \left( \mathbf{Z}_t \mathbf{R}_{t-1} \mathbf{Z}_t^\top \right) .
\end{align}

From which we derive the simplified inverse identity:

\begin{align}
\Bigl( \mathbf{I} + \mathbf{Z}_t \mathbf{R}_{t-1} \mathbf{Z}_t^\top \Bigr)^{-1} 
&= \mathbf{I} - 
\Bigl( \mathbf{I} + \mathbf{Z}_t \mathbf{R}_{t-1} \mathbf{Z}_t^\top \Bigr)^{-1} 
\Bigl( \mathbf{Z}_t \mathbf{R}_{t-1} \mathbf{Z}_t^\top \Bigr)
\label{eq:g}
\end{align}

Substituting Eq.~\eqref{eq:g} into the key expression within Eq.~\eqref{eq:f}, we obtain:

\begin{align}
& \mathbf{R}_{t-1} \mathbf{Z}_t^\top
\Bigl( \mathbf{I} + \mathbf{Z}_t \mathbf{R}_{t-1} \mathbf{Z}_t^\top \Bigr)^{-1} \\
&= \Bigl( \mathbf{R}_{t-1} - \mathbf{R}_{t-1} \mathbf{Z}_t^\top
\Bigl( \mathbf{I} + \mathbf{Z}_t \mathbf{R}_{t-1} \mathbf{Z}_t^\top \Bigr)^{-1}
\mathbf{Z}_t \mathbf{R}_{t-1} \Bigr) \mathbf{Z}_t^\top \\
&= \mathbf{R}_t \mathbf{Z}_t^\top
\label{eq:h}
\end{align}

Thus, plugging Eq.~\eqref{eq:h} into Eq.~\eqref{eq:f}, we simplify the update of the historical term as:

\begin{align}
\mathbf{R}_t \mathbf{Q}_{t-1} 
&= \mathbf{R}_{t-1} \mathbf{Q}_{t-1} -
\mathbf{R}_t \mathbf{Z}_t^\top
\mathbf{Z}_t \mathbf{R}_{t-1} \mathbf{Q}_{t-1} \\
&= \hat{\mathbf{W}}_{\mathrm{cls}}^{(t-1)} -
\mathbf{R}_t \mathbf{Z}_t^\top
\mathbf{Z}_t \hat{\mathbf{W}}_{\mathrm{cls}}^{(t-1)}
\label{eq:i}
\end{align}

This reveals a fundamental recursive structure: the new classifier is constructed from the old classifier by subtracting a correction term, which reflects the influence of the new session's input features on the prior solution.

\vspace{2mm}
\noindent
Finally, we substitute Eq.~\eqref{eq:i} into the classifier construction for session \( t \), yielding the complete recursive update formula:

\begin{align}
\hat{\mathbf{W}}_{\mathrm{cls}}^{(t)} 
= \left[
\hat{\mathbf{W}}_{\mathrm{cls}}^{(t-1)} -
\mathbf{R}_t \mathbf{Z}_t^\top
\mathbf{Z}_t \hat{\mathbf{W}}_{\mathrm{cls}}^{(t-1)} \quad \mathbf{R}_t \mathbf{Z}_t^\top \mathbf{Y}_t^{\mathrm{train}}
\right] .
\label{eq:recursive}
\end{align}

This confirms that Eq.~\eqref{eq:recursive} is equivalent to the original closed-form formulation Eq.~\eqref{eq:closed-from-solution}, and completes the recursive proof.

\subsection{Derivation of the Closed-form Solution for Regularized Least Squares}

We aim to solve the following regularized least squares optimization problem for obtaining the analytic classifier in the base stage of AL-GNN:

\begin{equation}
\min_{\hat{\mathbf{W}}_{\mathrm{cls}}^{(0)}} \left\| \mathbf{Y}_0^{\mathrm{train}} - \mathbf{Z}_0 \cdot \hat{\mathbf{W}}_{\mathrm{cls}}^{(0)} \right\|_F^2 + \gamma \left\| \hat{\mathbf{W}}_{\mathrm{cls}}^{(0)} \right\|_F^2 .
\label{eq:sup_obj}
\end{equation}

\vspace{1mm}
\noindent
Let us define the variables more compactly for the derivation:

\begin{itemize}
    \item Let \( \mathbf{Z} = \mathbf{Z}_0 \in \mathbb{R}^{n_0 \times d_{\mathrm{feg}}} \): expanded feature matrix with $n_0$ labeled nodes and $d_{\mathrm{feg}}$ dimensions.
    \item Let \( \mathbf{Y} = \mathbf{Y}_0^{\mathrm{train}} \in \mathbb{R}^{n_0 \times C_0} \): one-hot label matrix for the $C_0$ base classes.
    \item Let \( \mathbf{W} = \hat{\mathbf{W}}_{\mathrm{cls}}^{(0)} \in \mathbb{R}^{d_{\mathrm{feg}} \times C_0} \): analytic classifier weight matrix.
\end{itemize}

The objective function becomes:

\begin{equation}
\mathcal{L}(\mathbf{W}) = \| \mathbf{Y} - \mathbf{ZW} \|_F^2 + \gamma \| \mathbf{W} \|_F^2 .
\end{equation}

We now expand the Frobenius norm terms using the trace identity \( \| \mathbf{A} \|_F^2 = \mathrm{Tr}(\mathbf{A}^\top \mathbf{A}) \). Therefore:

\begin{align}
\mathcal{L}(\mathbf{W}) 
&= \mathrm{Tr}\left( (\mathbf{Y} - \mathbf{ZW})^\top (\mathbf{Y} - \mathbf{ZW}) \right) + \gamma \, \mathrm{Tr}(\mathbf{W}^\top \mathbf{W}) \\
&= \mathrm{Tr}(\mathbf{Y}^\top \mathbf{Y}) - 2 \mathrm{Tr}(\mathbf{Y}^\top \mathbf{ZW}) + \mathrm{Tr}(\mathbf{W}^\top \mathbf{Z}^\top \mathbf{Z} \mathbf{W}) + \gamma \, \mathrm{Tr}(\mathbf{W}^\top \mathbf{W}) .
\end{align}

Note that \( \mathrm{Tr}(\mathbf{Y}^\top \mathbf{Y}) \) is independent of \( \mathbf{W} \), and can be ignored when taking derivatives. Then the simplified objective is:

\begin{equation}
\mathcal{L}(\mathbf{W}) = -2 \mathrm{Tr}(\mathbf{Y}^\top \mathbf{ZW}) + \mathrm{Tr}(\mathbf{W}^\top (\mathbf{Z}^\top \mathbf{Z} + \gamma \mathbf{I}) \mathbf{W}) .
\end{equation}

We now take the derivative of \( \mathcal{L}(\mathbf{W}) \) with respect to \( \mathbf{W} \), using the following matrix calculus identities:

\begin{align}
\frac{\partial}{\partial \mathbf{W}} \mathrm{Tr}(\mathbf{A} \mathbf{W}) &= \mathbf{A}^\top , \\
\frac{\partial}{\partial \mathbf{W}} \mathrm{Tr}(\mathbf{W}^\top \mathbf{A} \mathbf{W}) &= 2 \mathbf{A} \mathbf{W} .
\end{align}

Applying them to our loss:

\begin{equation}
\frac{\partial \mathcal{L}}{\partial \mathbf{W}} = -2 \mathbf{Z}^\top \mathbf{Y} + 2 (\mathbf{Z}^\top \mathbf{Z} + \gamma \mathbf{I}) \mathbf{W} .
\end{equation}

Setting the gradient to zero:

\begin{equation}
-2 \mathbf{Z}^\top \mathbf{Y} + 2 (\mathbf{Z}^\top \mathbf{Z} + \gamma \mathbf{I}) \mathbf{W} = 0 .
\end{equation}

Dividing both sides by 2 and solving for \( \mathbf{W} \):

\begin{equation}
(\mathbf{Z}^\top \mathbf{Z} + \gamma \mathbf{I}) \mathbf{W} = \mathbf{Z}^\top \mathbf{Y} .
\end{equation}

\begin{equation}
\boxed{
\hat{\mathbf{W}}_{\mathrm{cls}}^{(0)} = (\mathbf{Z}_0^\top \mathbf{Z}_0 + \gamma \mathbf{I})^{-1} \mathbf{Z}_0^\top \mathbf{Y}_0^{\mathrm{train}} .
}
\end{equation}

This is the closed-form solution of multi-output ridge regression, which avoids backpropagation and enables fast, one-shot parameter computation in continual learning settings.

\subsection{Graph Convolutional Network (GCN) Formulation}

Given an undirected graph $\mathcal{G} = (\mathcal{V}, \mathcal{E})$ with $N = |\mathcal{V}|$ nodes and edges $\mathcal{E}$, we define its adjacency matrix as $\mathbf{A} \in \mathbb{R}^{N \times N}$ and its node feature matrix as $\mathbf{X} \in \mathbb{R}^{N \times d}$, where $d$ is the input feature dimension.

To allow each node to incorporate its own features during message passing, we augment the adjacency matrix by adding self-loops:
\begin{equation}
\tilde{\mathbf{A}} = \mathbf{A} + \mathbf{I},
\end{equation}
where $\mathbf{I}$ is the identity matrix.

The degree matrix corresponding to $\tilde{\mathbf{A}}$ is defined as:
\begin{equation}
\tilde{\mathbf{D}}_{ii} = \sum_j \tilde{\mathbf{A}}_{ij}.
\end{equation}

To ensure numerical stability and prevent exploding or vanishing gradients in deeper layers, we apply symmetric normalization to the adjacency matrix. The normalized adjacency matrix is given by:
\begin{equation}
\hat{\mathbf{A}} = \tilde{\mathbf{D}}^{-1/2} \tilde{\mathbf{A}} \tilde{\mathbf{D}}^{-1/2}.
\end{equation}

Now, let $\mathbf{H}^{(l)} \in \mathbb{R}^{N \times d_l}$ be the hidden representations of the nodes at the $l$-th layer, with $\mathbf{H}^{(0)} = \mathbf{X}$. A single-layer GCN performs the following propagation:
\begin{equation}
\mathbf{H}^{(l+1)} = \sigma \left( \hat{\mathbf{A}} \mathbf{H}^{(l)} \mathbf{W}^{(l)} \right),
\end{equation}
where $\mathbf{W}^{(l)} \in \mathbb{R}^{d_l \times d_{l+1}}$ is a learnable weight matrix, and $\sigma(\cdot)$ is a non-linear activation function, such as ReLU.

In the case of a two-layer GCN used for semi-supervised node classification, the model can be written as:
\begin{equation}
\hat{\mathbf{Y}} = \text{softmax} \left( \hat{\mathbf{A}} \cdot \sigma \left( \hat{\mathbf{A}} \cdot \mathbf{X} \cdot \mathbf{W}^{(0)} \right) \cdot \mathbf{W}^{(1)} \right),
\end{equation}
where $\hat{\mathbf{Y}} \in \mathbb{R}^{N \times C}$ denotes the predicted class probability distributions for $C$ classes.

The model is trained using the cross-entropy loss over the labeled nodes $\mathcal{V}_L \subset \mathcal{V}$:
\begin{equation}
\mathcal{L} = - \sum_{i \in \mathcal{V}_L} \sum_{c=1}^C Y_{ic} \log \hat{Y}_{ic},
\end{equation}
where $\mathbf{Y} \in \mathbb{R}^{N \times C}$ is the one-hot label matrix for the ground truth.

\paragraph{Interpretation.} 
This formulation can be interpreted as aggregating information from a node’s local neighborhood, applying a linear transformation, and then passing the result through a non-linear activation. The symmetric normalization ensures that nodes with high degrees do not overwhelm their neighbors during aggregation. By stacking multiple such layers, GCNs allow information to propagate across multiple hops in the graph, enabling the model to capture both feature and structural information.

In our work, GCN serves as the backbone feature extractor $f_{\mathrm{gcn}}(\cdot)$ to generate expressive node embeddings before applying the analytic classifier. Specifically, the GCN processes raw node features and aggregates local neighborhood information to produce intermediate representations $\mathbf{H}$. These representations are then passed through a fixed feature expansion layer and utilized in our analytic learning paradigm via closed-form ridge regression.

Unlike traditional end-to-end GCNs which rely entirely on backpropagation for training, AL-GNN decouples representation learning and classifier learning: the GCN is trained only in the base stage, and subsequent incremental updates are performed analytically without backpropagation. This design not only improves efficiency in continual learning scenarios, but also avoids catastrophic forgetting by preserving the GCN encoder and updating the classifier in a recursive, replay-free manner.

\clearpage
\section{Experiments Supplementary Material}
\subsection{Additional Experimental Results}
\begin{figure*}[htbp]
    \centering
    \includegraphics[width=1\textwidth]{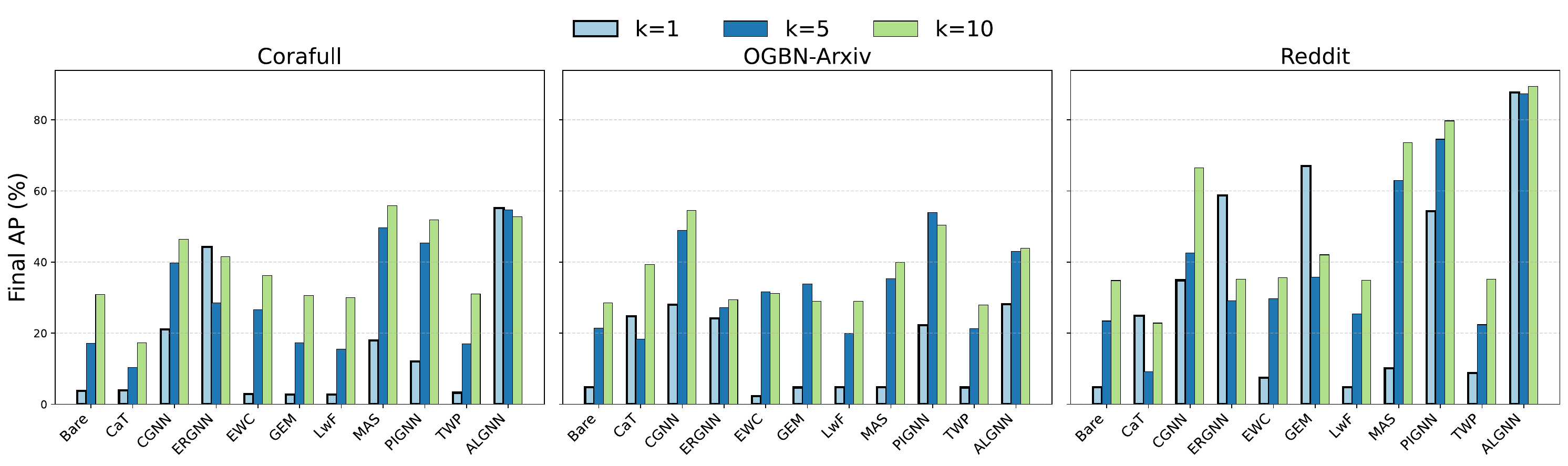}
    \caption{
    Final average precision (AP) comparison of 11 continual learning methods under class-incremental settings on three large-scale node classification datasets: \textbf{Corafull}, \textbf{OGBN-Arxiv}, and \textbf{Reddit}. 
    Each group of bars corresponds to one method, with three bars indicating different increment sizes \(k \in \{1, 5, 10\}\). 
    Smaller \(k\) represents more challenging class-incremental tasks with fewer new classes per step. 
    To highlight this, bars for \(k=1\) are shown with thicker outlines. 
    Overall, most methods exhibit improved performance with increasing \(k\), and \textbf{ALGNN} consistently outperforms most baselines across datasets and settings, demonstrating strong robustness and scalability.
    }
    \label{fig:final_ap_grouped_bar}
\end{figure*}

\begin{figure*}[htbp]
    \centering
    \includegraphics[width=1\textwidth]{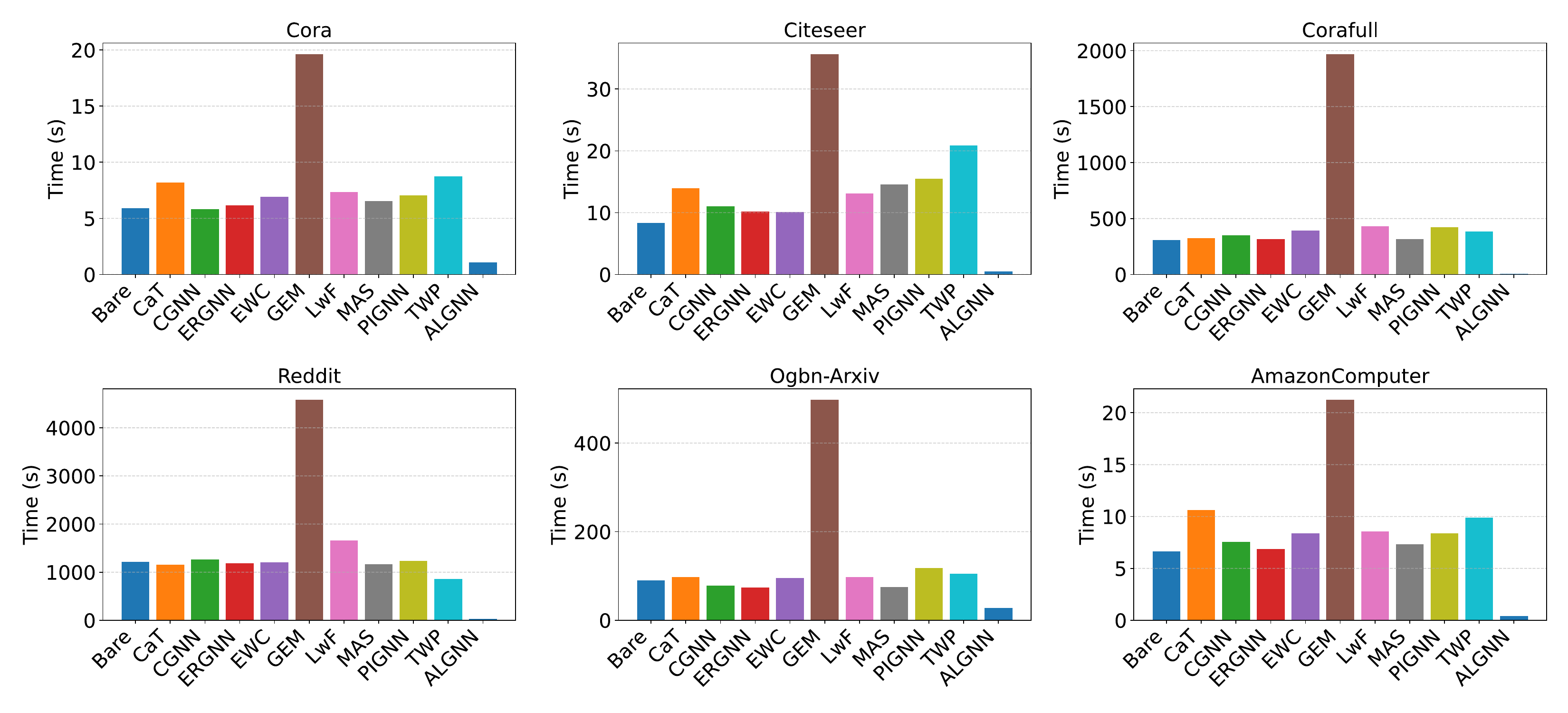}
    \caption{Grouped bar plots showing the total training time (in seconds) of 11 continual graph learning methods on six node classification datasets. The results reflect the cumulative time consumption including both the base stage and incremental updates. ALGNN consistently achieves the lowest or near-lowest time across all datasets.}

    \label{fig:time_by_dataset_all_algos_grouped.svg}
\end{figure*}

\begin{figure*}[htbp]
    \centering
    \includegraphics[width=1\textwidth]{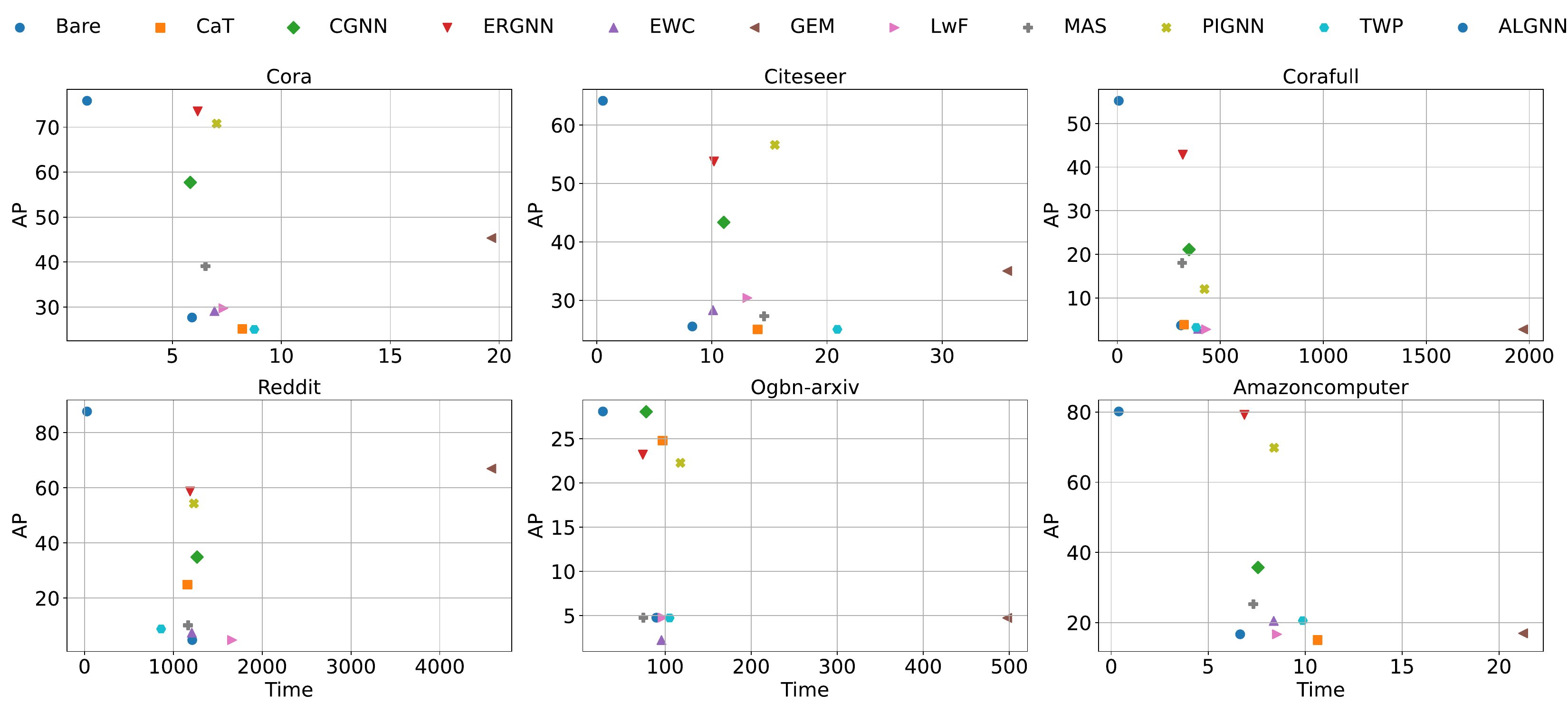}
    \caption{
    Comparison of average precision (AP) versus training time across six benchmark datasets: \textbf{Cora}, \textbf{Citeseer}, \textbf{Corafull}, \textbf{Reddit}, \textbf{OGBN-Arxiv}, and \textbf{AmazonComputer}. 
    Each scatter point corresponds to a continual graph learning method, using distinct markers and colors for visual differentiation. 
    The horizontal axis denotes the total training time (in seconds), while the vertical axis represents the final AP. 
    This figure reveals the trade-off between performance and efficiency, where methods located in the upper-left region demonstrate both high accuracy and low computational cost.
    Notably, \textbf{ALGNN} consistently achieves favorable trade-offs across all datasets, outperforming baselines in both effectiveness and efficiency.
    }\label{fig:ap_vs_time}
\end{figure*}

\clearpage
\subsection{Hyperparameter Settings}

We evaluate our method on six widely-used graph node classification benchmarks, covering both citation networks and large-scale real-world graphs. Cora and Citeseer are classical citation datasets with low-dimensional node features and a small number of classes, making them suitable for evaluating basic continual learning behavior. CoraFull extends Cora with a broader set of topics and 70 classes, posing greater challenges for class-incremental learning. Reddit is a large-scale social network with rich community structure, representing complex interactions and requiring scalable model inference. OGBN-Arxiv, from the Open Graph Benchmark, consists of ArXiv papers and captures temporal and structural evolution patterns, suitable for dynamic and evolving graphs. Lastly, Amazon-Computer is a co-purchase network with dense connections and higher feature dimensionality, reflecting realistic commercial recommendation scenarios. These datasets jointly enable a comprehensive evaluation across scale, sparsity, class diversity, and structural heterogeneity.
\begin{table}[htbp]
\centering
\caption{Basic statistics of the six node classification datasets used in our experiments.}
\label{tab:dataset_statistics}
\renewcommand{\arraystretch}{1.2}
\begin{tabular}{lccccc}
\toprule
\textbf{Dataset} & \textbf{\#Nodes} & \textbf{\#Edges} & \textbf{\#Classes} & \textbf{\#Features} & \textbf{Split Strategy} \\
\midrule
Cora              & 2,708   & 5,278    & 7    & 1,433 & Official split \\
Citeseer          & 3,327   & 4,732    & 6    & 3,703 & Official split \\
Corafull          & 19,793  & 63,421   & 70   & 871   & Random split (40/10/50) \\
Reddit            & 232,965 & 11,606,919 & 41 & 602   & Official split \\
OGBN-Arxiv        & 169,343 & 1,166,243 & 40   & 128   & OGB split \\
Amazon-Computer   & 13,752  & 245,861   & 10   & 767   & Random split (40/10/50) \\
\bottomrule
\end{tabular}
\end{table}

\begin{table}[htbp]
\centering
\caption{Hyperparameter settings for node classification experiments across different datasets and methods. ``Mem.'' denotes memory size (if applicable), ``$lr$'' denotes learning rate. ``--'' indicates not applicable.}
\label{tab:node_hyperparams}
\footnotesize
\setlength{\tabcolsep}{3pt}
\renewcommand{\arraystretch}{1.2}
\begin{tabular}{lccccccc}
\toprule
\textbf{Dataset} & \textbf{Tasks} & \textbf{Metric} & \textbf{Epochs} & \textbf{Batch} & \textbf{LR} & \textbf{Memory} & \textbf{Special Hyperparams (if any)} \\
\midrule
\textbf{Cora}            & 4  & Accuracy & 50 & 20 & 0.001 & 300  & See below \\
\textbf{Citeseer}        & 3  & Accuracy & 50 & 20 & 0.001 & 300  & See below \\
\textbf{Corafull}        & 33 & Accuracy & 50 & 20 & 0.001 & 500 & See below \\
\textbf{Reddit}          & 21 & Accuracy & 50 & 20 & 0.001 & 500 & See below \\
\textbf{OGBN-Arxiv}      & 21 & Accuracy & 50  & 50 & 0.001 & 300  & See below \\
\textbf{AmazonComputer}  & 6 & Accuracy & 50  & 20 & 0.001 & 200  & See below \\
\bottomrule
\end{tabular}

\vspace{1em}
\begin{tabular}{ll}
\toprule
\textbf{Method} & \textbf{Special Hyperparameters} \\
\midrule
Bare         & None \\
CGNN         & detect\_strategy=bfs, memory\_strategy=class, p=1, alpha=0.0, ewc\_lambda=0.5, ewc\_type=ewc \\
ERGNN        & sampler\_name=CM, distance\_threshold=0.5, num\_experience\_nodes=None \\
EWC          & $\lambda = 10000$ \\
GEM          & num\_memories as above, $\lambda = 0.5$ \\
LwF          & $\lambda = 1.0$, $T = 2.0$ \\
MAS          & $\lambda = 1.0$ \\
TWP          & $\lambda_t = 10.0$, $\lambda_l = 10.0$, $\beta = 0.01$ \\
CaT          & num\_memories as above \\
PIGNN        & retrain = 0.1 \\
\bottomrule
\end{tabular}
\end{table}

\clearpage
We divide the 7 classes of the Cora dataset into 4 sequential tasks and train a 2-layer GCN followed by a feature expansion module. Classifier weights are solved analytically using ridge regression and updated recursively. The experiment reports AP, AF, and training time with 3 independent runs.
\begin{table}[htbp]
\centering
\caption{Main experimental configuration for Cora node classification task.}
\label{tab:cora_hyperparam}
\renewcommand{\arraystretch}{1.2}
\begin{tabular}{ll}
\toprule
\textbf{Component} & \textbf{Setting} \\
\midrule
Dataset & Cora \\
Class Increment Setting & \([0,1,2,3] \rightarrow 4 \rightarrow 5 \rightarrow 6\) (4 tasks) \\
GCN Architecture & 2-layer: 1433 $\rightarrow$ 256 $\rightarrow$ 7 \\
Feature Expander & 1-layer GCN: 256 $\rightarrow$ 2048 \\
Dropout Rate & 0.5  \\
Optimizer & Adam \\
Learning Rate & 0.001 \\
Training Epochs (base stage) & 50 \\
Regularization Coefficient $\gamma$ & 1 \\
Classifier Update & Analytic solution + recursive update \\
Evaluation Metric & Average Precision (AP), Average Forgetting (AF), Time \\

\bottomrule
\end{tabular}
\end{table}

We adopt a 2-layer GCN model as the feature extractor followed by a single-layer feature expander. The classification is performed analytically using least-squares and recursively updated at each class-increment phase. The Citeseer dataset is split into four sequential tasks by class group. We measure AP, AF, and inference time for final evaluation.
\begin{table}[htbp]
\centering
\caption{Main experimental configuration for Citeseer node classification task.}
\label{tab:citeseer_hyperparam}
\renewcommand{\arraystretch}{1.2}
\begin{tabular}{ll}
\toprule
\textbf{Component} & \textbf{Setting} \\
\midrule
Dataset & Citeseer \\
Class Increment Setting & \([0,1,2] \rightarrow 3 \rightarrow 4 \rightarrow 5\) (4 tasks) \\
GCN Architecture & 2-layer: 3703 $\rightarrow$ 256 $\rightarrow$ 6 \\
Feature Expander & 1-layer GCN: 256 $\rightarrow$ 2048 \\
Dropout Rate & 0.5  \\
Optimizer & Adam \\
Learning Rate & 0.001 \\
Training Epochs (base stage) & 50 \\
Regularization Coefficient $\gamma$ & 1 \\
Classifier Update & Analytic solution + recursive update \\
Evaluation Metric & Average Precision (AP), Average Forgetting (AF), Time \\

\bottomrule
\end{tabular}
\end{table}

\begin{table}[htbp]
\centering
\caption{Main experimental configuration for CoraFull node classification under class-incremental setting.}
\label{tab:corafull_hyperparam}
\renewcommand{\arraystretch}{1.2}
\begin{tabular}{ll}
\toprule
\textbf{Component} & \textbf{Setting} \\
\midrule
Dataset & CoraFull \\
Total Classes & 70 \\
Class Increment Setting & 35 base classes, 35 incremental phases (1 class each) \\
Feature Extractor & 2-layer GCN: 871 $\rightarrow$ 256 $\rightarrow$ 35 \\
Dropout Rate & 0.5  \\
Feature Expander & 1-layer GCN: 256 $\rightarrow$ 2048 \\
Optimizer & Adam \\
Learning Rate & 0.001 \\
Epochs (Base Stage) & 50 \\
Regularization Coefficient ($\gamma$) & 0.001 \\
Classifier & Closed-form ridge regression with recursive update \\
Evaluation Metrics & AP, AF, Time \\
Device & CUDA \\

Train / Val / Test Split & 40\% / 10\% / 50\% (Random Seed 42) \\
\bottomrule
\end{tabular}
\end{table}

For Amazon-Computer, we adopt a 5+1×5 class-incremental learning setup, using a 2-layer GCN as feature extractor, followed by a one-layer expander. Classifier weights are computed via recursive ridge regression. The data is randomly split into training, validation, and test sets (40\%, 10\%, 50\%).

\begin{table}[htbp]
\centering
\caption{Main experimental configuration for Amazon-Computer node classification under class-incremental setting.}
\label{tab:amazon_hyperparam}
\renewcommand{\arraystretch}{1.2}
\begin{tabular}{ll}
\toprule
\textbf{Component} & \textbf{Setting} \\
\midrule
Dataset & Amazon Computer \\
Total Classes & 10 \\
Class Increment Setting & 5 base classes, 5 incremental phases (1 class each) \\
Feature Extractor & 2-layer GCN: 767 $\rightarrow$ 256 $\rightarrow$ 5 \\
Dropout Rate & 0.5  \\
Feature Expander & 1-layer GCN: 256 $\rightarrow$ 1024 \\
Optimizer & Adam \\
Learning Rate & 0.001 \\
Epochs (Base Stage) & 50 \\
Regularization Coefficient ($\gamma$) & 0.01 \\
Classifier & Closed-form ridge regression with recursive update \\
Evaluation Metrics & AP, AF, Time \\
Device & CUDA \\

Train / Val / Test Split & 40\% / 10\% / 50\% (Random Seed 42) \\
\bottomrule
\end{tabular}
\end{table}

\clearpage

\begin{table}[htbp]
\centering
\caption{Main experimental configuration for Reddit node classification under class-incremental setting.}
\label{tab:reddit_hyperparam}
\renewcommand{\arraystretch}{1.2}
\begin{tabular}{ll}
\toprule
\textbf{Component} & \textbf{Setting} \\
\midrule
Dataset & Reddit \\
Total Classes & 41 \\
Class Increment Setting & 21 base classes, 20 incremental phases (1 class each) \\
Feature Extractor & 2-layer GCN: 602 $\rightarrow$ 256 $\rightarrow$ 21 \\
Dropout Rate & 0.5 \\
Feature Expander & 1-layer GCN: 256 $\rightarrow$ 1024 \\
Optimizer & Adam \\
Learning Rate & 0.001 \\
Epochs (Base Stage) & 20 \\
Regularization Coefficient ($\gamma$) & 0.01 \\
Classifier & Closed-form ridge regression with recursive update \\
Evaluation Metrics & AP, AF, Time \\
Device & CUDA  \\

Train / Val / Test Split & Provided by dataset \\
\bottomrule
\end{tabular}
\end{table}

The OGBN-Arxiv dataset is split into one base phase (20 classes) and 20 incremental phases (each introducing a single class). A 2-layer GCN is used to extract features, followed by a high-dimensional graph-based expander. Classifier weights are computed using regularized least squares and updated recursively without backpropagation. There is a large gap in features between different classes on OGBN-Arxiv, and passing epochs will only make the effect worse.
\begin{table}[htbp]
\centering
\caption{Main experimental configuration for OGBN-Arxiv node classification with class-incremental setting.}
\label{tab:arxiv_hyperparam}
\renewcommand{\arraystretch}{1.2}
\begin{tabular}{ll}
\toprule
\textbf{Component} & \textbf{Setting} \\
\midrule
Dataset & OGBN-Arxiv \\
Total Classes & 40 \\
Class Increment Setting & First 20 as base, 20 incremental tasks with 1 class each \\
Feature Extractor & 2-layer GCN: 128 $\rightarrow$ 256 $\rightarrow$ 20 \\
Dropout Rate & 0.7  \\
Feature Expander & 1-layer GCN: 256 $\rightarrow$ 8192 \\
Optimizer & Adam \\
Learning Rate & 0.01 \\
Epochs (Base Stage) & 1 \\
Regularization Coefficient ($\gamma$) & 0.01 \\
Classifier & Closed-form ridge regression with recursive update \\
Evaluation Metrics & AP, AF, Time \\
Device & CUDA  \\

\bottomrule
\end{tabular}
\end{table}

\end{document}